%% file: _main.tex
\documentclass[10pt,letterpaper,twocolumn,pagenumbers]{article}
\usepackage[latin9]{inputenc}
\usepackage{float}
\usepackage{textcomp}
\usepackage{amsmath}
\usepackage{amsthm}
\usepackage{amssymb}
\usepackage{graphicx}
\usepackage{xargs}[2008/03/08]
\usepackage[unicode=true,
 bookmarks=false,
 breaklinks=true,pdfborder={0 0 1},backref=section,colorlinks=false]
 {hyperref}
\hypersetup{
 pagebackref=true,letterpaper=true,colorlinks}

\makeatletter

\pdfpageheight\paperheight
\pdfpagewidth\paperwidth

\providecommand{\tabularnewline}{\\}
\floatstyle{ruled}
\newfloat{algorithm}{tbp}{loa}
\providecommand{\algorithmname}{Algorithm}
\floatname{algorithm}{\protect\algorithmname}

\theoremstyle{plain}
\newtheorem{thm}{\protect\theoremname}
\ifx\proof\undefined
\newenvironment{proof}[1][\protect\proofname]{\par
	\normalfont\topsep6\p@\@plus6\p@\relax
	\trivlist
	\itemindent\parindent
	\item[\hskip\labelsep\scshape #1]\ignorespaces
}{%
	\endtrivlist\@endpefalse
}
\providecommand{\proofname}{Proof}
\fi
\newenvironment{lyxcode}
	{\par\begin{list}{}{
		\setlength{\rightmargin}{\leftmargin}
		\setlength{\listparindent}{0pt}% needed for AMS classes
		\raggedright
		\setlength{\itemsep}{0pt}
		\setlength{\parsep}{0pt}
		\normalfont\ttfamily}%
	 \item[]}
	{\end{list}}

\usepackage{iccv}
\usepackage{times}
\usepackage{epsfig}
\usepackage{graphicx}
\usepackage{amsmath}
\usepackage{amssymb}

\def\iccvPaperID{9343} % *** Enter the ICCV Paper ID here

\ificcvfinal\fi

\usepackage{cite}
\usepackage{amsmath,amssymb,amsfonts}
\usepackage{algorithmic}
\usepackage{graphicx}
\usepackage{textcomp}

\usepackage{amsthm}

\usepackage{algorithm}

\usepackage{enumitem}   
\@ifundefined{showcaptionsetup}{}{%
 \PassOptionsToPackage{caption=false}{subfig}}
\usepackage{subfig}
\makeatother

\providecommand{\theoremname}{Theorem}

\begin{document}
\title{A Class-aware Optimal Transport Approach with Higher-Order Moment
Matching for Unsupervised Domain Adaptation}
\author{Tuan Nguyen$^1$, Van Nguyen$^1$, Trung Le$^{1}$, He Zhao$^{2}$,
Quan Hung Tran$^3$, Dinh Phung$^{1}$\\
$^1$Department of Data Science and AI, Monash University, Australia\\
$^2$CSIRO\textquoteright s Data61,Australia\\
$^3$Adobe Research, San Jose, CA, USA\texttt{\small{}}~\\
\texttt{\small{}\{tuan.ng,}{\small{} }\texttt{\small{}van.nguyen1,
trunglm\}@monash.edu,he.zhao@ieee.org}{\small{}}\\
\texttt{\small{}qtran@adobe.com}{\small{},} \texttt{\small{}dinh.phung@monash.edu}}

\maketitle
\input{macros.tex}

\begin{abstract}
Unsupervised domain adaptation (UDA) aims to transfer knowledge from
a labeled source domain to an unlabeled target domain. In this paper,
we introduce a novel approach called class-aware optimal transport
(OT), which measures the OT distance between a distribution over the
source class-conditional distributions and a mixture of source and
target data distribution. Our class-aware OT leverages a cost function
that determines the matching extent between a given data example and
a source class-conditional distribution. By optimizing this cost function,
we find the optimal matching between target examples and source class-conditional
distributions, effectively addressing the data and label shifts that
occur between the two domains. To handle the class-aware OT efficiently,
we propose an amortization solution that employs deep neural networks
to formulate the transportation probabilities and the cost function.
Additionally, we propose minimizing class-aware Higher-order Moment
Matching (HMM) to align the corresponding class regions on the source
and target domains. The class-aware HMM component offers an economical
computational approach for accurately evaluating the HMM distance
between the two distributions. Extensive experiments on benchmark
datasets demonstrate that our proposed method significantly outperforms
existing state-of-the-art baselines. 
\end{abstract}

\section{Introduction}

\input{introduction.tex}

\section{Related Work}

\input{related_work.tex}

\section{Distributional OT Approach for Class-Aware UDA}

\input{our_approach.tex}

\section{High Order Moment Matching for UDA}

\input{hmm.tex}
\section{Experiment}

\input{experiment.tex}

\section{Conclusion}

In this paper, we present CLOTH, a novel method for unsupervised domain
adaptation (UDA). Our approach utilizes class-aware optimal transport
(OT) to measure the distance between a distribution over source class-conditional
distributions and a mixture distribution of source and target data.
To efficiently handle class-aware OT, we propose an amortization solution
that uses deep neural networks to amortize the transportation probabilities
and the cost function. In particular, we employ a multi-class source
discriminator as a deep network to amortize the cost function. Additionally,
we integrate class-aware Higher-order Moment Matching (HMM) into our
CLOTH framework, which combines both OT distance and HMM distance.
Extensive experiments demonstrate the superiority of our proposed
CLOTH over state-of-the-art baselines on benchmark datasets.

\appendix

\section{Background\label{sec:Background}}

\input{background.tex}

\section{All Proofs\label{sec:All_proofs}}

In this section, we present the proofs for Theorem \ref{thm:amortize}.
\begin{proof}
Let $\left(G^{*},A^{*}=\left[a_{im}^{*}\right]_{i,m},\bpi^{*}\right)$
be the solution of the optimization problem (OP) in (2). Due to the
infinite capacity of the family of the transportation networks, there
exists $\mathcal{T}^{*},G^{*}$ such that $\mathcal{T}^{*}\left(G^{*}\left(\bx_{i}\right)\right)=N\ba_{i}^{*},\forall i$
where $\ba_{i}^{*}=\left[a_{im}^{*}\right]_{i}$.

For any transportation network $\mathcal{T}$ and feature extractor
$G$, we denote $\ba_{i}=\frac{1}{N}\mathcal{T}\left(G\left(\bx_{i}\right)\right),\forall i$,
$A=\left[\ba_{1}\,\ba_{2}\,...\,\ba_{N}\right]^{T}$, and $\bpi=\left[\pi_{m}\right]_{m}$
with $\pi_{m}=\sum_{i=1}^{N}a_{im}$. Since $\left(G^{*},A^{*},\bpi^{*}\right)$
is the optimal solution of the OP in (2), we have:

\[
\sum_{i=1}^{N}\sum_{m=1}^{M}a_{im}^{*}c\left(G^{*}\left(\bx_{i}\right),\mathbb{Q}_{m}^{S}\left(G^{*}\right)\right)
\]
\begin{equation}
\leq\sum_{i=1}^{N}\sum_{m=1}^{M}a_{im}c\left(G\left(\bx_{i}\right),\mathbb{Q}_{m}^{S}\left(G\right)\right),
\end{equation}
where $\mathbb{Q}_{m}^{S}\left(G^{*}\right)$ and $\mathbb{Q}_{m}^{S}\left(G\right)$
represent the corresponding distributions on the latent space w.r.t.
$G^{*}$and $G$.
\[
\sum_{i=1}^{N}\sum_{m=1}^{M}\mathcal{T}_{m}^{*}\left(G^{*}\left(\bx_{i}\right)\right)c\left(G^{*}\left(\bx_{i}\right),\mathbb{Q}_{m}^{S}\left(G^{*}\right)\right)
\]
\begin{equation}
\leq\sum_{m=1}^{M}\mathcal{T}_{m}\left(G\left(\bx_{i}\right)\right)c\left(G\left(\bx_{i}\right),\mathbb{Q}_{m}^{S}\left(G\right)\right).
\end{equation}

Therefore, $\left(G^{*},\mathcal{T}^{*}\right)$ is the optimal solution
of the OP in (3).

Let $\left(G^{*},\mathcal{T}^{*}\right)$ be the optimal solution
of the OP in (3). We denote $\ba_{i}^{*}=\frac{1}{N}\mathcal{T}^{*}\left(G^{*}\left(\bx_{i}\right)\right),\forall i$,
$A=\left[\ba_{1}^{*}\,\ba_{2}^{*}\,...\,\ba_{N}^{*}\right]^{T}$,
and $\bpi^{*}=\left[\pi_{m}^{*}\right]_{m}$ with $\pi_{m}^{*}=\sum_{i=1}^{N}a_{im}^{*}$.
Let $\left(G,A=\left[a_{im}\right]_{i,m},\bpi\right)$ be a feasible
solution of the OP in (2). Due to the infinite capacity of the family
of the transportation networks, there exists $\mathcal{T},G$ such
that $\mathcal{T}\left(G\left(\bx_{i}\right)\right)=N\ba_{i},\forall i$
where $\ba_{i}=\left[a_{im}\right]_{i}$. Since $\left(G^{*},\mathcal{T}^{*}\right)$
is the optimal solution of the OP in (3), we have:

\[
\sum_{i=1}^{N}\sum_{m=1}^{M}\mathcal{T}_{m}^{*}\left(G^{*}\left(\bx_{i}\right)\right)c\left(G^{*}\left(\bx_{i}\right),\mathbb{Q}_{m}^{S}\left(G^{*}\right)\right)
\]
\begin{equation}
\leq\sum_{m=1}^{M}\mathcal{T}_{m}\left(G\left(\bx_{i}\right)\right)c\left(G\left(\bx_{i}\right),\mathbb{Q}_{m}^{S}\left(G\right)\right).
\end{equation}

\[
\sum_{i=1}^{N}\sum_{m=1}^{M}a_{im}^{*}c\left(G^{*}\left(\bx_{i}\right),\mathbb{Q}_{m}^{S}\left(G^{*}\right)\right)
\]
\begin{equation}
\leq\sum_{i=1}^{N}\sum_{m=1}^{M}a_{im}c\left(G\left(\bx_{i}\right),\mathbb{Q}_{m}^{S}\left(G\right)\right).
\end{equation}

Therefore, $\left(G^{*},A^{*}=\left[a_{im}^{*}\right]_{i,m},\bpi^{*}\right)$
is the solution of the OP in (2). 
\end{proof}

\section{Network Architecture}

\label{sec:Net_arch}

In the experiments on the \emph{Digits}, inspired by HoMM \cite{chen2020homm},
we utilize a modified LeNet \cite{lecun1998lenet} architecture for
the generator $G$. This modified architecture includes a dense layer
with $90$ hidden neurons inserted before the final output dense layers
of the classifier $\mathcal{C}$, discriminator $D$, and transportation
$\mathcal{T}$. Due to the varying sizes of images across different
domains, we resize all images to $32\times32$ before using them as
inputs for our models. Additionally, RGB images are converted to grayscale.

In the experiments on the \emph{Office-31}, \emph{Office-Home}, and
\emph{ImageCLEF-DA} datasets, we employ pre-extracted features with
a dimension of $2048$ from ResNet-50 \cite{he2016resnet}. ResNet-50
is a widely used deep learning model known for its effectiveness in
feature extraction and has been utilized in various state-of-the-art
methods such as HoMM \cite{chen2020homm}, SPL \cite{WangSPL2019},
and SHOT \cite{liang2020shot}. By leveraging the extracted ResNet-50
features, we only need to design a simple network for the generator
$G$. For the \emph{Office-Home} and \emph{ImageCLEF-DA} datasets,
the generator consists of a single dense layer with $256$ hidden
neurons. For the \emph{Office-31} dataset, it includes two hidden
dense layers with $1024$ and $90$ neurons, respectively. These layers
are followed by the final output dense layers for the classifier $\mathcal{C}$,
discriminator $D$, and transportation $\mathcal{T}$ networks. Additionally,
we incorporate Dropout layers with a fixed retention probability of
either $0.5$ or $0.8$ as a form of regularization in the generator
$G$, classifier $\mathcal{C}$, discriminator $D$, and transportation
$\mathcal{T}$ networks.

\section{Additional Analyses}

\label{sec:Add_anal}

In the subsequent sections, we conduct a range of ablation studies
to gain deeper insights into our model. These studies encompass the
following aspects:

(i) In-depth analysis of factors that influence the model's performance,
including the effect of Class-aware Higher-Order Moment Matching and
the contributions of adversarial training.

(ii) Investigation of the model's behavior by comparing the current
approach and architecture with alternative methods. For instance,
we compare the Sinkhorn algorithm \cite{cuturi2013sinkhorn} with
our amortization solution, and explore the performance when the transportation
$\mathcal{T}$ and classifier $\mathcal{C}$ share weights and when
they have separate weights.

(iii) Assessment of the model's sensitivity to changes in hyperparameters,
such as $\alpha$, $\beta$, and $\gamma$.

(iv) Presentation of qualitative results, including \emph{t}-SNE visualization,
to showcase the performance of our proposed CLOTH.

\subsection{Comparison between the baseline HoMM and CaHoMM}

We conduct a comparison between our proposed CaHoMM and HoMM \cite{chen2020homm}
using two criteria: classification accuracy and training time. We
set up two settings for this comparison: \emph{CLOT$+$HoMM} and \emph{CLOTH}.

In the \emph{CLOT$+$HoMM} setting, we use all component losses as
described in (22) but replace $\mathcal{L}^{HMM}$ with the \emph{arbitrary-order
moment matching} loss from the authors' source code\footnote{https://github.com/chenchao666/HoMM-Master}.

In the \emph{CLOTH} setting, we apply our proposed method with CaHoMM.
To ensure a fair comparison, we keep the same settings, including
the architecture and hyperparameters, for both \emph{CLOT$+$HoMM}
and \emph{CLOTH}.

These two experiments are executed on an NVIDIA Tesla V100 SXM2 with
16 GB memory. 
\begin{itemize}
\item \textbf{Classification accuracy.} The results are reported in Table
\ref{tab:CAHoMM_effect}. Our proposed \emph{CLOTH} (the fourth row)
achieves an average accuracy of $91.5\%$, outperforming \emph{CLOT$+$HoMM}
(the third row) by $1\%$. 
\item \textbf{Training comparison.} Table \ref{tab:training_comparison}
presents the training time for the two settings on transfer tasks
\textbf{A}$\rightarrow$\textbf{W} and \textbf{P}$\rightarrow$\textbf{I}.
Both \emph{CLOT$+$HoMM} and \emph{CLOTH} have the same number of
trainable parameters, but our proposed model shows significant improvements
in training time. For the \textbf{A}$\rightarrow$\textbf{W} task,
\emph{CLOTH} is $44.49\%$ faster in training time per batch and $34.25\%$
faster in total training time compared to \emph{CLOT$+$HoMM}. Similarly,
for the \textbf{P}$\rightarrow$\textbf{I} task, \emph{CLOTH} has
a training time per batch of only $0.06353$ seconds ($44.03\%$ faster)
and a total training time of $1650.52$ seconds ($34.08\%$ faster)
compared to \emph{CLOT$+$HoMM}. These results demonstrate the efficiency
of our proposed economical formulation for computing the HMM distance,
resulting in significant reductions in space complexity from $\mathcal{O}\left(p^{q}\right)$
to $\mathcal{O}\left(p\right)$ and time complexity by performing
simple vector dot-product computations (see (\ref{eq:L_HMM})). 
\end{itemize}
\begin{table}[H]
\caption{Training comparison of our proposed CLOTH with and without CaHoMM
on two transfer tasks: \textbf{A}$\rightarrow$\textbf{W} (Office-31)
and \textbf{P}$\rightarrow$\textbf{I} (ImageCLEF-DA). \label{tab:training_comparison}}
\centering{}\resizebox{1.0\columnwidth}{!}{%%%
\begin{tabular}{cccc}
\hline 
Method & \# Parameters & Time/batch (s) & Total time (s)\tabularnewline
\hline 
CLOT$+$HoMM (\textbf{A}$\rightarrow$\textbf{W}) & 2199104 & 2.79704 & 56262.49\tabularnewline
CLOTH (\textbf{A}$\rightarrow$\textbf{W}) & 2199104 & 0.06287 & 1642.59\tabularnewline
\hline 
CLOT$+$HoMM (\textbf{P}$\rightarrow$\textbf{I}) & 2193841 & 2.79697 & 56252.84\tabularnewline
CLOTH (\textbf{P}$\rightarrow$\textbf{I}) & 2193841 & 0.06353 & 1650.52\tabularnewline
\hline 
\end{tabular}}
\end{table}

\subsection{Effect of Adversarial Training Framework}

To investigate the impact of adversarial training (AT) on the model
performance, we propose an experiment with two settings: \emph{CLOTH$-$AT}
and \emph{CLOTH$+$AT}.

In \emph{CLOTH$-$AT}, we ignore AT during training, and the objective
function becomes minimizing $\mathcal{L}^{C}+\alpha\mathcal{L}^{t}+\beta\mathcal{L}^{ent}$
with respect to $\mathcal{C}$, $G$, and $\mathcal{T}$ (The loss
$\mathcal{L}^{HMM}$ is not used in both settings for a fair comparison).

In \emph{CLOTH$+$AT}, we incorporate AT into our CLOTH model. The
loss function in this setting is similar to the one in (12), where
we minimize $\mathcal{L}^{C}+\mathcal{L}^{G,S}+\mathcal{L}^{G,T}+\alpha\mathcal{L}^{t}+\beta\mathcal{L}^{ent}$
with respect to $\mathcal{C}$, $G$, $\mathcal{T}$, and alternatively
update $\mathcal{D}$ to minimize $\mathcal{L}^{\mathcal{D}}$.

Table \ref{tab:Abla_AT_officehome} shows the results of this experiment
on the \emph{Office-Home} dataset. The \emph{CLOTH$+$AT} setting
significantly outperforms \emph{CLOTH$-$AT} by $2.4\%$. This improvement
can be attributed to the effectiveness of AT, which helps mix up source
and target samples in a class-aware manner and enables the multi-class
discriminator to produce accurate outputs for computing the cost $c\left(G\left(\bx_{i}\right),\mathbb{Q}_{m}^{S}\right)$.
Consequently, the transportation network $\mathcal{T}$ can learn
more accurately and approximate the optimal transportation matrix
$A^{*}$ in (3).

\begin{table*}
\caption{Results (\%) of our proposed CLOTH with and without adversarial training
on Office-Home.\label{tab:Abla_AT_officehome}}
\centering{}\resizebox{1.0\textwidth}{!}{%%%
\begin{tabular}{cccccccccccccc}
\hline 
Method & Ar$\rightarrow$Cl & Ar$\rightarrow$Pr & Ar$\rightarrow$Re & Cl$\rightarrow$Ar & Cl$\rightarrow$Pr & Cl$\rightarrow$Re & Pr$\rightarrow$Ar & Pr$\rightarrow$Cl & Pr$\rightarrow$Re & Re$\rightarrow$Ar & Re$\rightarrow$Cl & Re$\rightarrow$Pr & Avg\tabularnewline
\hline 
CLOTH$-$AT & 51.5 & 75.6 & 80.9 & \textbf{66.6} & 78.4 & 77.6 & 65.2 & 51.9 & 79.8 & 70.4 & 53.7 & 83.4 & 69.6\tabularnewline
CLOTH$+$AT & \textbf{57.2} & \textbf{78.4} & \textbf{82.6} & 66.1 & \textbf{80.2} & \textbf{81.2} & \textbf{65.6} & \textbf{55.1} & \textbf{82.8} & \textbf{71.6} & \textbf{59.2} & \textbf{83.9} & \textbf{72.0}\tabularnewline
\hline 
\end{tabular}}
\end{table*}

\subsection{Performance Comparison: Sinkhorn Algorithm vs. Our Amortization Solution}

One interesting study is the comparison between our amortization solution
and the Sinkhorn algorithm \cite{cuturi2013sinkhorn} in solving the
optimization problem in (3). We first rewrite the optimization problem
in (3) in the form of an entropic regularized version:

\begin{align}
\mathcal{W}_{c,\bpi}^{\epsilon}\left(\mathbb{Q},\mathcal{Q}^{S}\right) & =\min_{A}\biggl\{\sum_{i=1}^{N}\sum_{m=1}^{M}a_{im}c\left(G\left(\bx_{i}\right),\mathbb{Q}_{m}^{S}\right)\nonumber \\
-\epsilon H(A): & \sum_{m=1}^{M}a_{im}=\frac{1}{N_{T}},\sum_{i=1}^{N}a_{im}=\pi_{m}\biggr\},\label{eq:ws_latent_entropic}
\end{align}

%where the cost to transport a data sample $G\left(\bx_{i}\right)$
%to $\mathbb{Q}_{m}^{S}$ is defined as $c\left(G\left(\bx_{i}\right),\mathbb{Q}_{m}^{S}\right)$$=-\log\mathcal{D}_{m}\left(G\left(\bx_{i}\right)\right)$,
%$H(A)\coloneqq-\sum_{i=1}^{N}\sum_{m=1}^{M}a_{im}\log a_{im}$ denotes an entropy of the transportation matrix $A$, and $\epsilon=0.1$
is the regularization rate. During the training, we solve this OP
using the Sinkhorn algorithm and achieve $A^{*}$ at every mini-batch.
We now design two optimization problems to fairly compare two approaches: 
\begin{itemize}
\item To solve the optimization problem in (3) using the Sinkhorn algorithm,
the final objective function is defined as: 
\end{itemize}
\begin{equation}
\min_{\mathcal{C},G,\mathcal{T}}\left\{ \mathcal{L}^{C}+\mathcal{L}^{G,S}+\mathcal{L}^{G,T}+\alpha\mathcal{W}_{c,\bpi}^{\epsilon}\left(\mathbb{Q},\mathcal{Q}^{S}\right)\right\} 
\end{equation}

\begin{itemize}
\item To solve the optimization problem in (3) using our amortization solution,
we formulate the final objective function, which resembles the one
in (12):
\end{itemize}
\begin{equation}
\min_{\mathcal{C},G,\mathcal{T}}\left\{ \mathcal{L}^{C}+\mathcal{L}^{G,S}+\mathcal{L}^{G,T}+\alpha\mathcal{L}^{t}+\beta\mathcal{L}^{ent}\right\} 
\end{equation}

In both scenarios, we update alternatively $\mathcal{D}$ by minimizing
the loss function $\mathcal{L}^{\mathcal{D}}$. Two experiments were
conducted on the \emph{Office-31} dataset, and the results are reported
in Table \ref{tab:Sinkhorn_vs_Armortized}. The accuracy scores obtained
using the Sinkhorn algorithm (the \emph{Sinkhorn} setting) are significantly
lower compared to our approach (the \emph{Amortization} setting).
These results highlight the difficulty of the Sinkhorn algorithm in
accurately approximating the Wasserstein distance between the empirical
mixture distribution $\mathbb{Q}$ of the source and target domains
in the latent space and the source class-conditional distributions
$\mathbb{Q}_{m}^{S}$. This is attributed to the limited batch size,
which makes it challenging to precisely approximate the optimal transportation
matrix $A^{*}$. Consequently, pushing target samples to the desired
$\mathbb{Q}_{m}^{S}$ distributions becomes challenging, leading to
a decline in model performance. On the other hand, our approach (\emph{Amortization})
significantly outperforms the \emph{Sinkhorn} setting by a large margin
(24.2\%). This demonstrates the effectiveness of approximating $A^{*}$
using the transportation network $\mathcal{T}$ trained on mini-batches.
The transportation network $\mathcal{T}$ produces accurate transportation
probabilities for each target sample, enabling them to be effectively
transported to the appropriate source class regions.

\begin{table}[H]
\centering{}\caption{Results (\%) of our proposed CLOTH with two different optimal transport
strategies on Office-31.\label{tab:Sinkhorn_vs_Armortized}}
\resizebox{1.00\columnwidth}{!}{%%%
\begin{tabular}{cccccccc}
\hline 
Method & A$\rightarrow$W & A$\rightarrow$D & D$\rightarrow$W & W$\rightarrow$D & D$\rightarrow$A & W$\rightarrow$A & Avg\tabularnewline
\hline 
Sinkhorn & 53.2 & 47.8 & 82.5 & 79.9 & 75.7 & 72.5 & 68.6\tabularnewline
Amortization & \textbf{95.6} & \textbf{95.0} & \textbf{98.1} & \textbf{100.0} & \textbf{87.9} & \textbf{80.0} & \textbf{92.8}\tabularnewline
\hline 
\end{tabular}}
\end{table}

\subsection{Hyper-parameter Sensitivity}

\begin{figure*}
\begin{centering}
\includegraphics[width=0.3\textwidth]{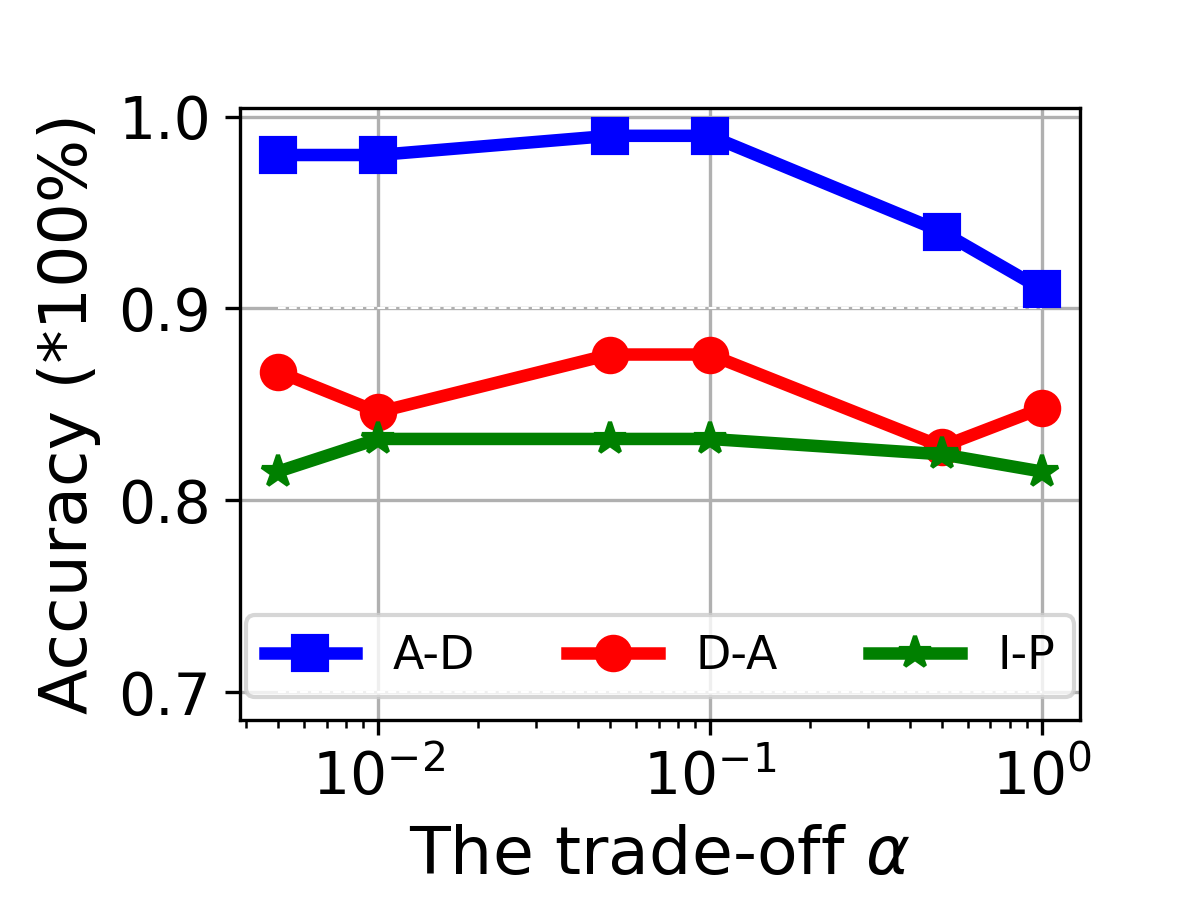}\hfill{}\includegraphics[width=0.3\textwidth]{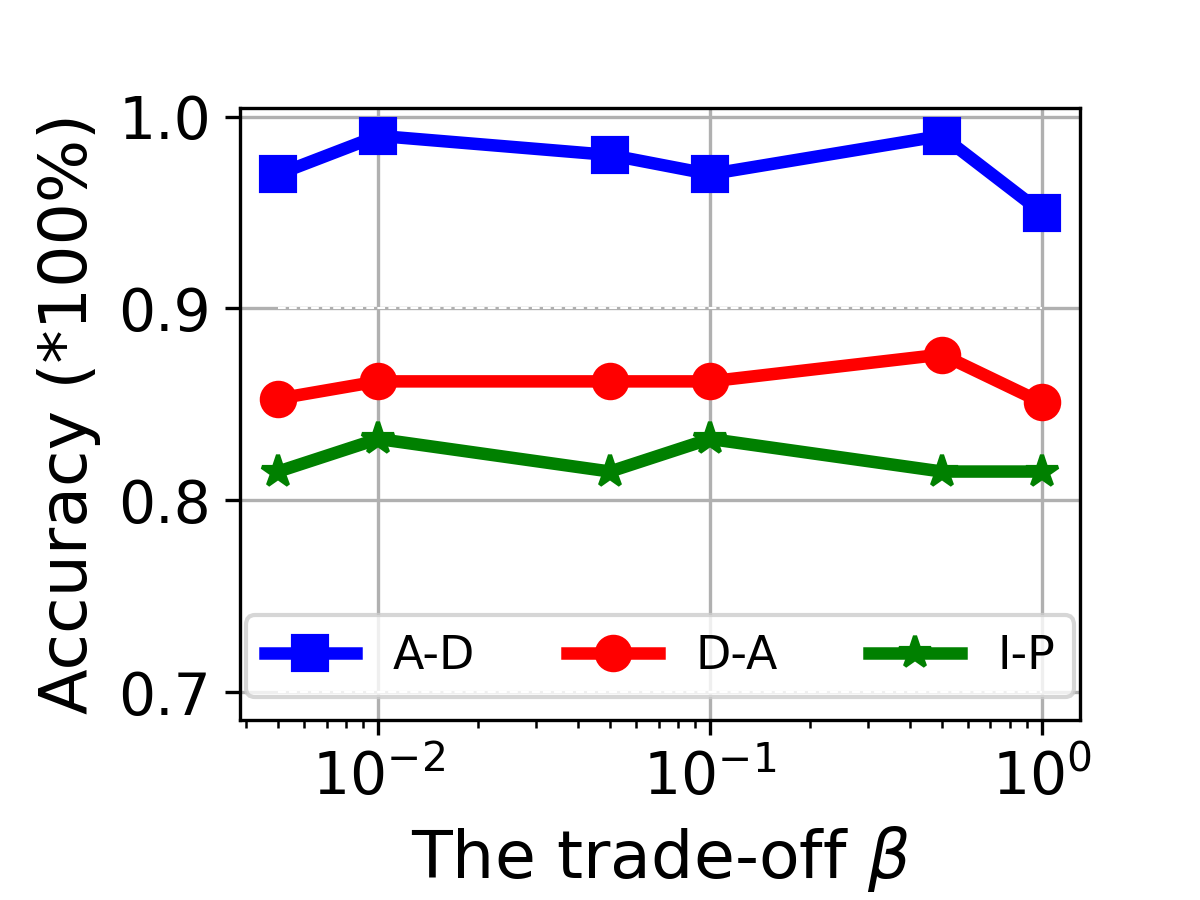}\hfill{}\includegraphics[width=0.3\textwidth]{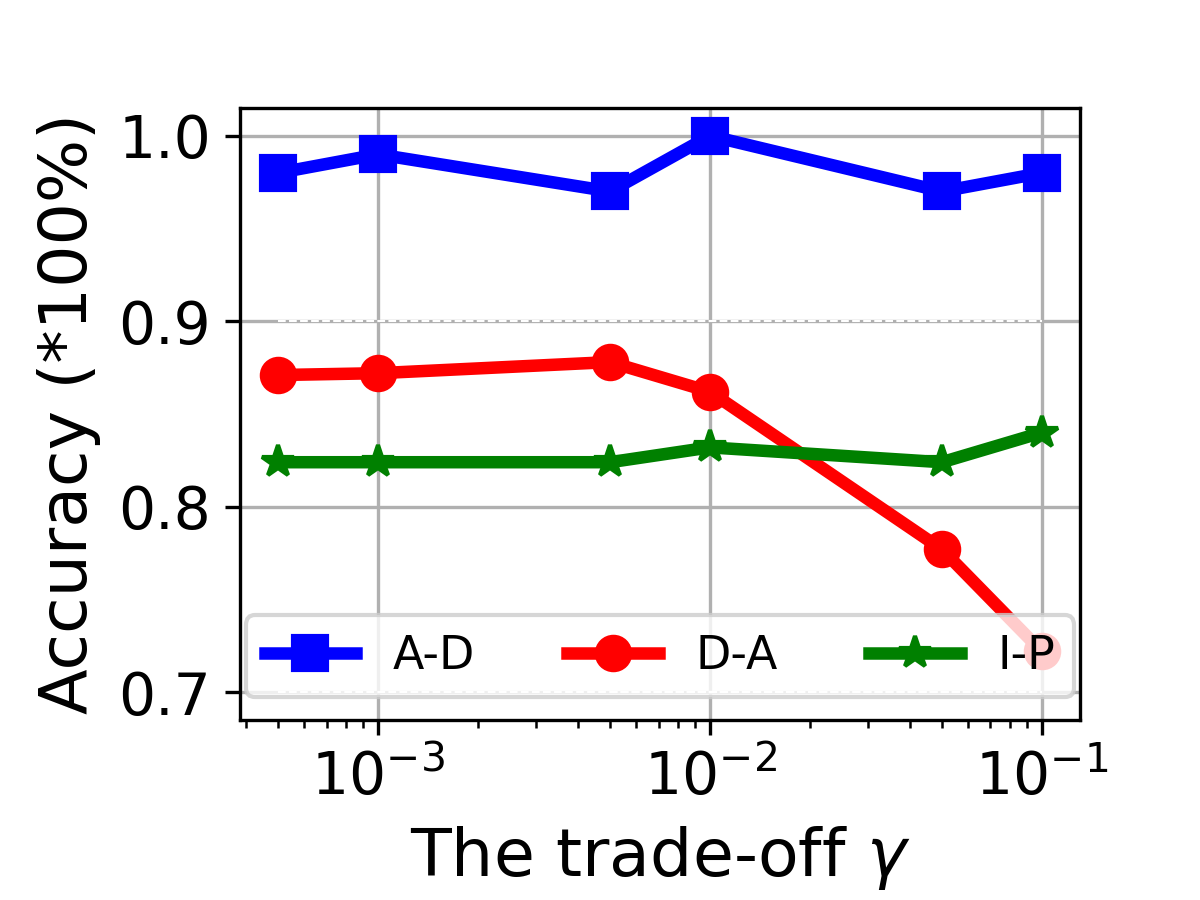}\vspace{-1mm}
\par\end{centering}
\caption{Analysis of hyper-parameter sensitivity of $\alpha,\beta$ and $\gamma$
on transfer tasks \textbf{A}$\rightarrow$\textbf{D}, \textbf{D}$\rightarrow$\textbf{A}
(Office-31) and \textbf{I}$\rightarrow$\textbf{P} (ImageCLEF-DA).\label{fig:abla_hypersen}}
\vspace{-1mm}
\end{figure*}

We also conducted experiments to investigate the sensitivity of our
model's performance to the hyperparameters, specifically the values
of $\alpha$, $\beta$, and $\gamma$, within the recommended ranges.
The test accuracy scores on three transfer tasks, \textbf{A}$\rightarrow$\textbf{D},
\textbf{D}$\rightarrow$\textbf{A}, and \textbf{I}$\rightarrow$\textbf{P},
are presented in Figure \ref{fig:abla_hypersen}. The results indicate
that our CLOTH consistently achieves stable performances when $\alpha$
and $\beta$ are set within the range of $\left\{ 10^{-2},10^{-1}\right\} $,
and $\gamma$ is set within the range of $\left\{ 10^{-3},10^{-2}\right\} $.
Based on empirical observations, we find that most transfer tasks
achieve good performances when $\alpha=\beta=10^{-1}$ and $\gamma=10^{-2}$.

\subsection{Shared and Unshared $\mathcal{T}$ and $\mathcal{C}$}

In terms of modeling, we considered whether it is advantageous to
share the weights of the classifier network $\mathcal{C}$ and the
transportation network $\mathcal{T}$. We investigated two cases:
(i) shared weights between $\mathcal{C}$ and $\mathcal{T}$ (\emph{S\_CT}),
and (ii) unshared weights between $\mathcal{C}$ and $\mathcal{T}$
(\emph{U\_CT}). The experimental results presented in Table \ref{tab:The-test-accuracy-office-31-1}
demonstrate that when $\mathcal{T}$ differs from $\mathcal{C}$,
our proposed model achieves higher accuracy performance in most transfer
tasks (\textbf{A}$\rightarrow$\textbf{W}, \textbf{W}$\rightarrow$\textbf{D},
\textbf{D}$\rightarrow$\textbf{A}, \textbf{W}$\rightarrow$\textbf{A})
compared to the case where $\mathcal{T}$ and $\mathcal{C}$ are identical.
Based on this analysis, we choose to construct separate networks $\mathcal{C}$
and $\mathcal{T}$ in all our experiments.

\begin{table}
\centering{}\caption{Results (\%) of our proposed CLOTH with different model architectures
for $\mathcal{C}$ and $\mathcal{T}$ on Office-31.\label{tab:The-test-accuracy-office-31-1}}
\resizebox{1.00\columnwidth}{!}{%%%
\begin{tabular}{cccccccc}
\hline 
Method & A$\rightarrow$W & A$\rightarrow$D & D$\rightarrow$W & W$\rightarrow$D & D$\rightarrow$A & W$\rightarrow$A & Avg\tabularnewline
\hline 
S\_CT & 95.2 & \textbf{96.0} & \textbf{98.5} & 99.8 & 87.3 & 84.4 & 93.5\tabularnewline
U\_CT & \textbf{96.9} & \textbf{96.0} & 98.1 & \textbf{100.0} & \textbf{87.4} & \textbf{85.8} & \textbf{94.0}\tabularnewline
\hline 
\end{tabular}}
\end{table}

\subsection{Feature Visualization}

\begin{figure*}
\begin{centering}
\includegraphics[width=0.6\textwidth]{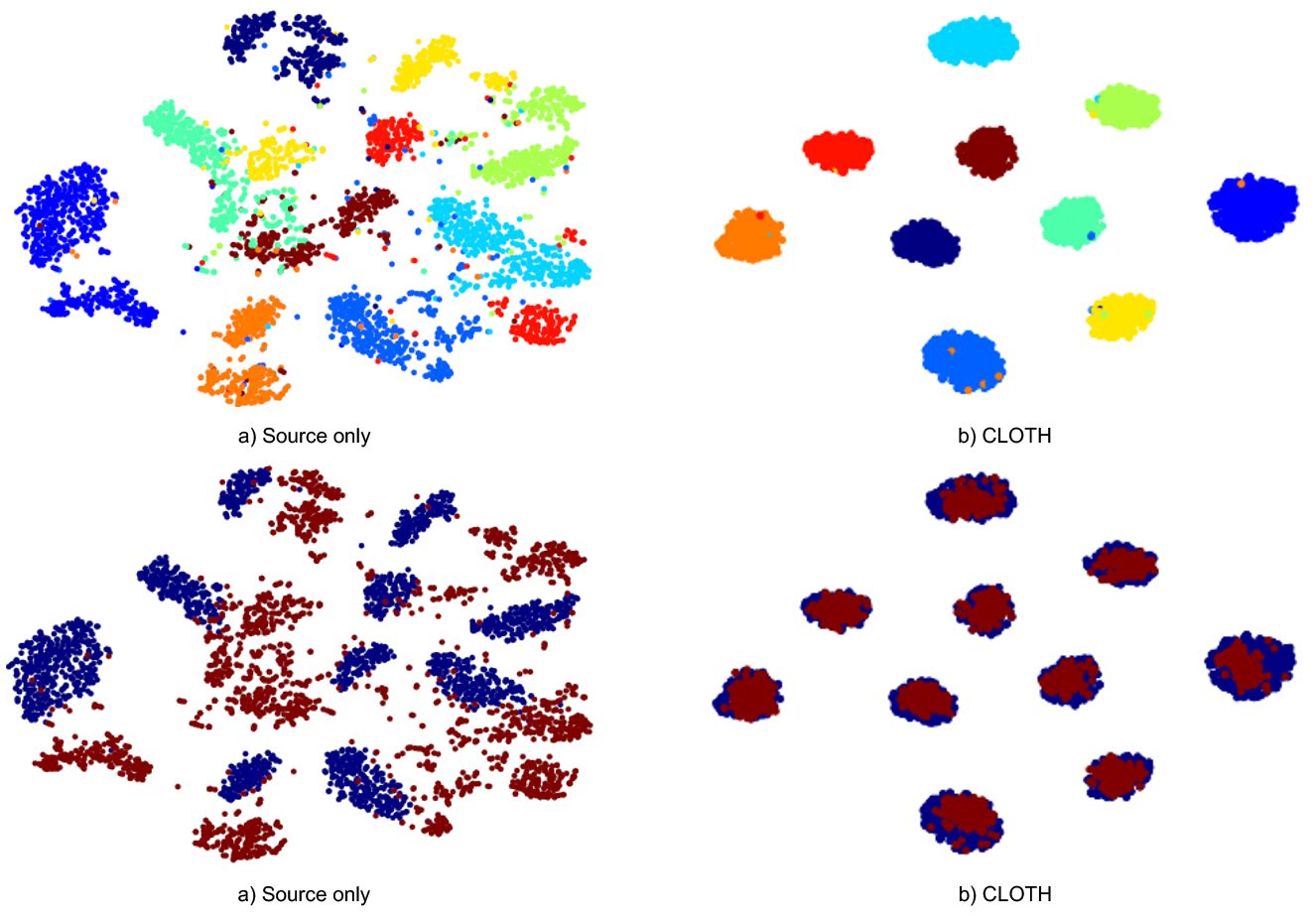}
\par\end{centering}
\caption{The \emph{t}-SNE visualization of the transfer task from SVHN to MNIST
in the latent space. The first row shows the features labeled by category
information, where each color represents a category. The second row
visualizes these features labeled by domain information, with blue
and red points representing the source and target samples, respectively.\label{fig:The-t-SNE-visualization}}
\end{figure*}

In this section, we further demonstrate the effectiveness of our proposed
CLOTH in terms of feature transferability. We use a \emph{t}-SNE \cite{vandermaaten2008visualizing}
projection to visualize the feature distributions of the source and
target data in the joint space. Specifically, we apply this visualization
to the transfer task from \textbf{SVHN} to \textbf{MNIST}. The results,
shown in Figure \ref{fig:The-t-SNE-visualization}, illustrate that
the feature representations generated by our CLOTH exhibit clear boundaries
and form exactly 10 clusters, which correspond to the 10 classes of
\emph{Digits}. These visualizations demonstrate that our method is
capable of aligning the complex structures of the source and target
data, while effectively maximizing the margin between different classes.
This success is attributed to the guidance provided by the transportation
network $\mathcal{T}$, the effectiveness of adversarial training
with the multi-discriminator $\mathcal{D}$, and the integration of
class-aware higher-order moment matching in our approach.

\section*{Acknowledgment}

This work was supported by the US Air Force grant FA2386-19-1-4040.
\begin{lyxcode}
\clearpage{}
\end{lyxcode}
\bibliographystyle{ieee_fullname}
\bibliography{eccv22}

\end{document}

%% file: macros.tex
\global\long\def\sidenote#1{\marginpar{\small\emph{{\color{Medium}#1}}}}%

\global\long\def\se{\hat{\text{se}}}%
\global\long\def\interior{\text{int}}%
\global\long\def\boundary{\text{bd}}%
\global\long\def\ML{\textsf{ML}}%
\global\long\def\GML{\mathsf{GML}}%
\global\long\def\HMM{\mathsf{HMM}}%
\global\long\def\support{\text{supp}}%
\global\long\def\new{\text{*}}%
\global\long\def\stir{\text{Stirl}}%
\global\long\def\mA{\mathcal{A}}%
\global\long\def\mB{\mathcal{B}}%
\global\long\def\expect{\mathbb{E}}%
\global\long\def\mF{\mathcal{F}}%
\global\long\def\mK{\mathcal{K}}%
\global\long\def\mH{\mathcal{H}}%
\global\long\def\mX{\mathcal{X}}%
\global\long\def\mZ{\mathcal{Z}}%
\global\long\def\mS{\mathcal{S}}%
\global\long\def\Ical{\mathcal{I}}%
\global\long\def\mT{\mathcal{T}}%
\global\long\def\Pcal{\mathcal{P}}%
\global\long\def\dist{d}%
\global\long\def\HX{\entro\left(X\right)}%
\global\long\def\entropyX{\HX}%
\global\long\def\HY{\entro\left(Y\right)}%
\global\long\def\entropyY{\HY}%
\global\long\def\HXY{\entro\left(X,Y\right)}%
\global\long\def\entropyXY{\HXY}%
\global\long\def\mutualXY{\mutual\left(X;Y\right)}%
\global\long\def\mutinfoXY{\mutualXY}%
\global\long\def\given{\mid}%
\global\long\def\gv{\given}%
\global\long\def\goto{\rightarrow}%
\global\long\def\asgoto{\stackrel{a.s.}{\longrightarrow}}%
\global\long\def\pgoto{\stackrel{p}{\longrightarrow}}%
\global\long\def\dgoto{\stackrel{d}{\longrightarrow}}%
\global\long\def\lik{\mathcal{L}}%
\global\long\def\logll{\mathit{l}}%
\global\long\def\bigcdot{\raisebox{-0.5ex}{\scalebox{1.5}{\ensuremath{\cdot}}}}%
\global\long\def\sig{\textrm{sig}}%
\global\long\def\likelihood{\mathcal{L}}%
\global\long\def\vectorize#1{\mathbf{#1}}%

\global\long\def\vt#1{\mathbf{#1}}%
\global\long\def\gvt#1{\boldsymbol{#1}}%
\global\long\def\idp{\ \bot\negthickspace\negthickspace\bot\ }%
\global\long\def\cdp{\idp}%
\global\long\def\das{}%
\global\long\def\id{\mathbb{I}}%
\global\long\def\idarg#1#2{\id\left\{  #1,#2\right\}  }%
\global\long\def\iid{\stackrel{\text{iid}}{\sim}}%
\global\long\def\bzero{\vt 0}%
\global\long\def\bone{\mathbf{1}}%
\global\long\def\a{\mathrm{a}}%
\global\long\def\ba{\mathbf{a}}%
\global\long\def\b{\mathrm{b}}%
\global\long\def\bb{\mathbf{b}}%
\global\long\def\B{\mathrm{B}}%
\global\long\def\boldm{\boldsymbol{m}}%
\global\long\def\c{\mathrm{c}}%
\global\long\def\C{\mathrm{C}}%
\global\long\def\d{\mathrm{d}}%
\global\long\def\D{\mathrm{D}}%
\global\long\def\N{\mathrm{N}}%
\global\long\def\h{\mathrm{h}}%
\global\long\def\H{\mathrm{H}}%
\global\long\def\bH{\mathbf{H}}%
\global\long\def\K{\mathrm{K}}%
\global\long\def\M{\mathrm{M}}%
\global\long\def\bff{\vt f}%
\global\long\def\bx{\mathbf{\mathbf{x}}}%

\global\long\def\bl{\boldsymbol{l}}%
\global\long\def\s{\mathrm{s}}%
\global\long\def\T{\mathrm{T}}%
\global\long\def\bu{\mathbf{u}}%
\global\long\def\v{\mathrm{v}}%
\global\long\def\bv{\mathbf{v}}%
\global\long\def\bo{\boldsymbol{o}}%
\global\long\def\bh{\mathbf{h}}%
\global\long\def\bs{\boldsymbol{s}}%
\global\long\def\x{\mathrm{x}}%
\global\long\def\bx{\mathbf{x}}%
\global\long\def\bz{\mathbf{z}}%
\global\long\def\hbz{\hat{\bz}}%
\global\long\def\z{\mathrm{z}}%
\global\long\def\y{\mathrm{y}}%
\global\long\def\bxnew{\boldsymbol{y}}%
\global\long\def\bX{\boldsymbol{X}}%
\global\long\def\tbx{\tilde{\bx}}%
\global\long\def\by{\mathbf{y}}%
\global\long\def\bY{\boldsymbol{Y}}%
\global\long\def\bZ{\boldsymbol{Z}}%
\global\long\def\bU{\boldsymbol{U}}%
\global\long\def\bn{\boldsymbol{n}}%
\global\long\def\bV{\boldsymbol{V}}%
\global\long\def\bI{\boldsymbol{I}}%
\global\long\def\J{\mathrm{J}}%
\global\long\def\bJ{\mathbf{J}}%
\global\long\def\w{\mathrm{w}}%
\global\long\def\bw{\vt w}%
\global\long\def\bW{\mathbf{W}}%
\global\long\def\balpha{\gvt{\alpha}}%
\global\long\def\bdelta{\boldsymbol{\delta}}%
\global\long\def\bsigma{\gvt{\sigma}}%
\global\long\def\bbeta{\gvt{\beta}}%
\global\long\def\bmu{\gvt{\mu}}%
\global\long\def\btheta{\boldsymbol{\theta}}%
\global\long\def\blambda{\boldsymbol{\lambda}}%
\global\long\def\bgamma{\boldsymbol{\gamma}}%
\global\long\def\bpsi{\boldsymbol{\psi}}%
\global\long\def\bphi{\boldsymbol{\phi}}%
\global\long\def\bpi{\boldsymbol{\pi}}%
\global\long\def\bomega{\boldsymbol{\omega}}%
\global\long\def\bepsilon{\boldsymbol{\epsilon}}%
\global\long\def\btau{\boldsymbol{\tau}}%
\global\long\def\bxi{\boldsymbol{\xi}}%
\global\long\def\realset{\mathbb{R}}%
\global\long\def\realn{\realset^{n}}%
\global\long\def\integerset{\mathbb{Z}}%
\global\long\def\natset{\integerset}%
\global\long\def\integer{\integerset}%

\global\long\def\natn{\natset^{n}}%
\global\long\def\rational{\mathbb{Q}}%
\global\long\def\rationaln{\rational^{n}}%
\global\long\def\complexset{\mathbb{C}}%
\global\long\def\comp{\complexset}%

\global\long\def\compl#1{#1^{\text{c}}}%
\global\long\def\and{\cap}%
\global\long\def\compn{\comp^{n}}%
\global\long\def\comb#1#2{\left({#1\atop #2}\right) }%
\global\long\def\param{\vt w}%
\global\long\def\Param{\Theta}%
\global\long\def\meanparam{\gvt{\mu}}%
\global\long\def\Meanparam{\mathcal{M}}%
\global\long\def\meanmap{\mathbf{m}}%
\global\long\def\logpart{A}%
\global\long\def\simplex{\Delta}%
\global\long\def\simplexn{\simplex^{n}}%
\global\long\def\dirproc{\text{DP}}%
\global\long\def\ggproc{\text{GG}}%
\global\long\def\DP{\text{DP}}%
\global\long\def\ndp{\text{nDP}}%
\global\long\def\hdp{\text{HDP}}%
\global\long\def\gempdf{\text{GEM}}%
\global\long\def\rfs{\text{RFS}}%
\global\long\def\bernrfs{\text{BernoulliRFS}}%
\global\long\def\poissrfs{\text{PoissonRFS}}%
\global\long\def\grad{\gradient}%
\global\long\def\gradient{\nabla}%
\global\long\def\partdev#1#2{\partialdev{#1}{#2}}%
\global\long\def\partialdev#1#2{\frac{\partial#1}{\partial#2}}%
\global\long\def\partddev#1#2{\partialdevdev{#1}{#2}}%
\global\long\def\partialdevdev#1#2{\frac{\partial^{2}#1}{\partial#2\partial#2^{\top}}}%
\global\long\def\closure{\text{cl}}%
\global\long\def\cpr#1#2{\Pr\left(#1\ |\ #2\right)}%
\global\long\def\var{\text{Var}}%
\global\long\def\Var#1{\text{Var}\left[#1\right]}%
\global\long\def\cov{\text{Cov}}%
\global\long\def\Cov#1{\cov\left[ #1 \right]}%
\global\long\def\COV#1#2{\underset{#2}{\cov}\left[ #1 \right]}%
\global\long\def\corr{\text{Corr}}%
\global\long\def\sst{\text{T}}%
\global\long\def\SST{\sst}%
\global\long\def\ess{\mathbb{E}}%

\global\long\def\Ess#1{\ess\left[#1\right]}%
\newcommandx\ESS[2][usedefault, addprefix=\global, 1=]{\underset{#2}{\ess}\left[#1\right]}%
\global\long\def\fisher{\mathcal{F}}%

\global\long\def\bfield{\mathcal{B}}%
\global\long\def\borel{\mathcal{B}}%
\global\long\def\bernpdf{\text{Bernoulli}}%
\global\long\def\betapdf{\text{Beta}}%
\global\long\def\dirpdf{\text{Dir}}%
\global\long\def\gammapdf{\text{Gamma}}%
\global\long\def\gaussden#1#2{\text{Normal}\left(#1, #2 \right) }%
\global\long\def\gauss{\mathbf{N}}%
\global\long\def\gausspdf#1#2#3{\text{Normal}\left( #1 \lcabra{#2, #3}\right) }%
\global\long\def\multpdf{\text{Mult}}%
\global\long\def\poiss{\text{Pois}}%
\global\long\def\poissonpdf{\text{Poisson}}%
\global\long\def\pgpdf{\text{PG}}%
\global\long\def\wshpdf{\text{Wish}}%
\global\long\def\iwshpdf{\text{InvWish}}%
\global\long\def\nwpdf{\text{NW}}%
\global\long\def\niwpdf{\text{NIW}}%
\global\long\def\studentpdf{\text{Student}}%
\global\long\def\unipdf{\text{Uni}}%
\global\long\def\transp#1{\transpose{#1}}%
\global\long\def\transpose#1{#1^{\mathsf{T}}}%
\global\long\def\mgt{\succ}%
\global\long\def\mge{\succeq}%
\global\long\def\idenmat{\mathbf{I}}%
\global\long\def\trace{\mathrm{tr}}%
\global\long\def\argmax#1{\underset{_{#1}}{\text{argmax}} }%
\global\long\def\argmin#1{\underset{_{#1}}{\text{argmin}\ } }%
\global\long\def\diag{\text{diag}}%
\global\long\def\norm{}%
\global\long\def\spn{\text{span}}%
\global\long\def\vtspace{\mathcal{V}}%
\global\long\def\field{\mathcal{F}}%
\global\long\def\ffield{\mathcal{F}}%
\global\long\def\inner#1#2{\left\langle #1,#2\right\rangle }%
\global\long\def\iprod#1#2{\inner{#1}{#2}}%
\global\long\def\dprod#1#2{#1 \cdot#2}%
\global\long\def\norm#1{\left\Vert #1\right\Vert }%
\global\long\def\entro{\mathbb{H}}%
\global\long\def\entropy{\mathbb{H}}%
\global\long\def\Entro#1{\entro\left[#1\right]}%
\global\long\def\Entropy#1{\Entro{#1}}%
\global\long\def\mutinfo{\mathbb{I}}%
\global\long\def\relH{\mathit{D}}%
\global\long\def\reldiv#1#2{\relH\left(#1||#2\right)}%
\global\long\def\KL{KL}%
\global\long\def\KLdiv#1#2{\KL\left(#1\parallel#2\right)}%
\global\long\def\KLdivergence#1#2{\KL\left(#1\ \parallel\ #2\right)}%
\global\long\def\crossH{\mathcal{C}}%
\global\long\def\crossentropy{\mathcal{C}}%
\global\long\def\crossHxy#1#2{\crossentropy\left(#1\parallel#2\right)}%
\global\long\def\breg{\text{BD}}%
\global\long\def\lcabra#1{\left|#1\right.}%
\global\long\def\lbra#1{\lcabra{#1}}%
\global\long\def\rcabra#1{\left.#1\right|}%
\global\long\def\rbra#1{\rcabra{#1}}%

%% file: introduction.tex
Unsupervised domain adaptation (UDA) allows us to transfer knowledge
from a model trained on a source domain with labels to a target domain
without any labels. To cope more efficiently and effectively with
structural data, deep domain adaptation (DDA) \cite{Ganin2015} has
been proposed and extensively studied. Additionally, to address the
data shift issue and learn domain-invariant features, DDA aims to
bridge the distribution gap between the source and target domains
in a latent space using a feature extractor. Guided by this principle,
most existing works in DDA propose minimizing the divergence between
the source and target distributions in the latent space. Popular choices
of divergence include the Jensen-Shannon (JS) divergence \cite{Ganin2015,TzengHDS15,shu2018a},
the maximum mean discrepancy (MMD) distance \cite{gretton2007kernel,long2015},
and the Wasserstein (WS) distance \cite{shen2018wasserstein,chenyu2019swd}.

Recently, optimal transport (OT) \cite{santambrogio2015optimal,villani2008optimal},
a discipline in mathematics with a rich and rigorous theory, has been
widely applied in deep learning, particularly in domain adaptation
\cite{courty2017optimal,courty2017joint,damodaran2018deepjdot,RedkoCFT19,chenyu2019swd,yujia2019onscalable,xu2020reliable}.
From a conceptual perspective, the OT-based objective function encourages
the target examples to move to the source examples by minimizing a
transportation cost. However, since the transportation cost generally
considers pairs of target and source examples without taking into
account the label information of the source examples, the movement
of the target examples to the source domain seems to be unaware of
the class regions in that domain, thus unable to resolve the label
shift issue. Although OT has been initially used to address this problem
\cite{courty2017optimal,damodaran2018deepjdot}, the performance of
the existing methods is still less satisfactory compared to state-of-the-art
approaches.

In this paper, we propose a novel class-aware optimal transport approach
that incorporates source label information when matching target and
source examples. Specifically, we consider a distribution of distributions
where each component distribution represents a class-conditional distribution
of the source domain. We then compute an OT distance between this
distribution and the mixture of target and source data distribution,
where the cost function determines the matching degree between an
example and a class-conditional distribution. By minimizing this OT
distance, we aim to find the optimal matching between target examples
and source class-conditional distributions, effectively incorporating
source label information. To handle the OT distance, we propose an
amortization solution that utilizes deep neural networks to amortize
transportation probabilities and the cost function. Additionally,
inspired by the work of \cite{chen2020homm}, we introduce \textbf{\textit{C}}\textit{lass-}\textbf{\textit{a}}\textit{ware
}\textbf{\textit{H}}\textit{igher-}\textbf{\textit{o}}\textit{rder
}\textbf{\textit{M}}\textit{oment }\textbf{\textit{M}}\textit{atching}
(CaHoMM) distance to enhance the matching efficiency between corresponding
class regions in the source and target domains. We derive an efficient
and precise formulation to evaluate this HMM distance, making it a
lightweight component that can be incorporated into our class-aware
approach. We propose a method called \textbf{\textit{CL}}\textit{ass-aware
}\textbf{\textit{O}}\textit{ptimal }\textbf{\textit{T}}\textit{ransport
with }\textbf{\textit{H}}\textit{igher-Order Moment matching} (CLOTH).

In summary, our contributions in this paper are as follows:
\begin{itemize}
\item We propose a novel class-aware optimal transport method that addresses
the challenges of data and label shift in Unsupervised Domain Adaptation
(UDA). Our approach considers source label information and seeks the
optimal matching between target and source examples to mitigate these
issues. 
\item We introduce a Class-aware Higher-order Moment Matching (CaHoMM) distance
to accurately approximate the distributions of the source and target
domains in the latent space. This CaHoMM distance is evaluated using
an elegant and efficient formulation, improving the class-aware matching
between the two domains. 
\item We conduct extensive experiments to compare our proposed method, CLOTH,
with state-of-the-art baselines in standard UDA, class-aware UDA,
and OT-based UDA on benchmark datasets such as Digits, Office-31,
Office-Home, and ImageCLEF-DA. Our proposed CLOTH achieves state-of-the-art
performance on these benchmark datasets, surpassing existing methods. 
\end{itemize}
Overall, our contributions include the development of a class-aware
OT method, the introduction of class-aware Higher-Order Moment Matching,
and the empirical validation of our approach's superior performance
on various benchmark datasets.

%% file: related_work.tex
\subsection{Standard DA}

Deep domain adaptation (DA) has received significant attention and
has demonstrated impressive performance in various tasks and applications,
as highlighted in previous studies \cite{Ganin2015,long2015,saito2017asymmetric,french2018selfensembling}.
The fundamental concept behind deep DA is to minimize the distributional
gap between the source and target domains in a shared feature space
by reducing the divergence between the distributions induced by these
domains. Commonly used divergence measures include Jensen-Shannon
divergence \cite{Ganin2015,TzengHDS15,shu2018a}; maximum mean discrepancy
distance \cite{gretton2007kernel,long2015}; and Wasserstein distance
\cite{shen2018wasserstein,chenyu2019swd}. Recent research has explored
different aspects of unsupervised domain adaptation (UDA) to enhance
performance, such as adversarial adaptation \cite{KurmiCADA2019,pmlr-v97-chen19i,awais2021RFA},
labeling and propagation \cite{WangSPL2019,zhang2020label}, transferability
and feature alignment \cite{Wang2019CDAN,liang2020shot,chen2020homm,le2021lamda}.

\subsection{Optimal Transport based DA}

OT theory has been applied to DA in \cite{courty2017optimal,courty2017joint,damodaran2018deepjdot,RedkoCFT19,chenyu2019swd,yujia2019onscalable,xu2020reliable,MengxueETD2020,le2021lamda}.
Particularly, \cite{chenyu2019swd} proposed using sliced-Wasserstein
distance for DA, whereas \cite{yujia2019onscalable} proposed SPOT,
in which the OT plan is approximated by a pushforward of a reference
distribution. Courty et al. \cite{courty2017optimal} proposed an
idea to connect the theory of OT and DA \cite{courty2017joint}, which
later inspired an OT-based deep DA method (DeepJDOT) \cite{damodaran2018deepjdot}.
Recent OT-based DA work (RWOT) \cite{xu2020reliable} leveraged spatial
prototypical information and intra-domain structures of image data
to reduce the negative transfer caused by target samples near decision
boundaries. \cite{MengxueETD2020} proposed ETD to measure the domain
discrepancy under the guidance of the prediction-feedback via developing
a novel attention-aware OT distance, while LAMDA \cite{le2021lamda}
has been proposed to measure the label shift guaranteed by OT theory
and identify the disadvantages of learning domain-invariant representations.
Different from existing OT-based approaches, our proposed method examines
an OT distance between a distribution over source class-conditional
distributions and a mixture of source and target data distribution.
By investigating this specific OT distance and solving it using an
amortization approach, we can guide target examples to move toward
an appropriate source class in the latent space to mitigate both data
and label shifts.

\subsection{Class-aware DA}

Some recent approaches, such as \cite{Kang2019classaware,Wang2019classaware},
have leveraged useful information from the label space to improve
the quality of alignment between the source and target domains. Wang
et al. \cite{Wang2019classaware} proposed a novel relationship-aware
adversarial domain adaptation (RADA) algorithm that uses a single
multi-class domain discriminator to enforce the learning of the inter-class
dependency structure during domain-adversarial training. RADA aligns
this structure with the inter-class dependencies characterized from
training the label predictor on the source domain, making the adversarial
domain adaptation aware of the class relationships. Kang et al. \cite{Kang2019classaware}
introduced a contrastive adaptation network (CAN) that optimizes a
new metric modeling both the intra-class domain discrepancy and the
inter-class domain discrepancy, enabling class-aware unsupervised
domain adaptation (UDA).

%% file: our_approach.tex
\subsection{Problem Formulation}

We consider the vanilla setting of unsupervised domain adaptation
in which we have a labeled dataset $\mathbb{D}^{S}=\left\{ \left(\bx_{i}^{S},y_{i}^{S}\right)\right\} _{i=1}^{N_{S}}$
from a source domain and another unlabeled dataset $\mathbb{D}^{T}=\left\{ \bx_{i}^{T}\right\} _{i=1}^{N_{T}}$
from a target domain. We assume that data examples $\bx_{i}^{S},\bx_{i}^{T}\in\mathbb{R}^{d}$
and the categorical labels $y_{i}^{S}\in\left\{ 1,2,...,M\right\} $
where $M$ is the number of classes. We denote $\mathbb{P}^{S}$ and
$\mathbb{P}^{T}$ as the data distributions of the source and target
domains, respectively. Moreover, given a class $m$, we further denote
$\mathbb{P}_{m}^{S}$ as the $m$-th class-conditional distribution
of the source domain (i.e., the distribution with the density function
$p^{S}\left(\bx\mid y=m\right)$).

\begin{figure}
\centering{\includegraphics[width=0.6\columnwidth]{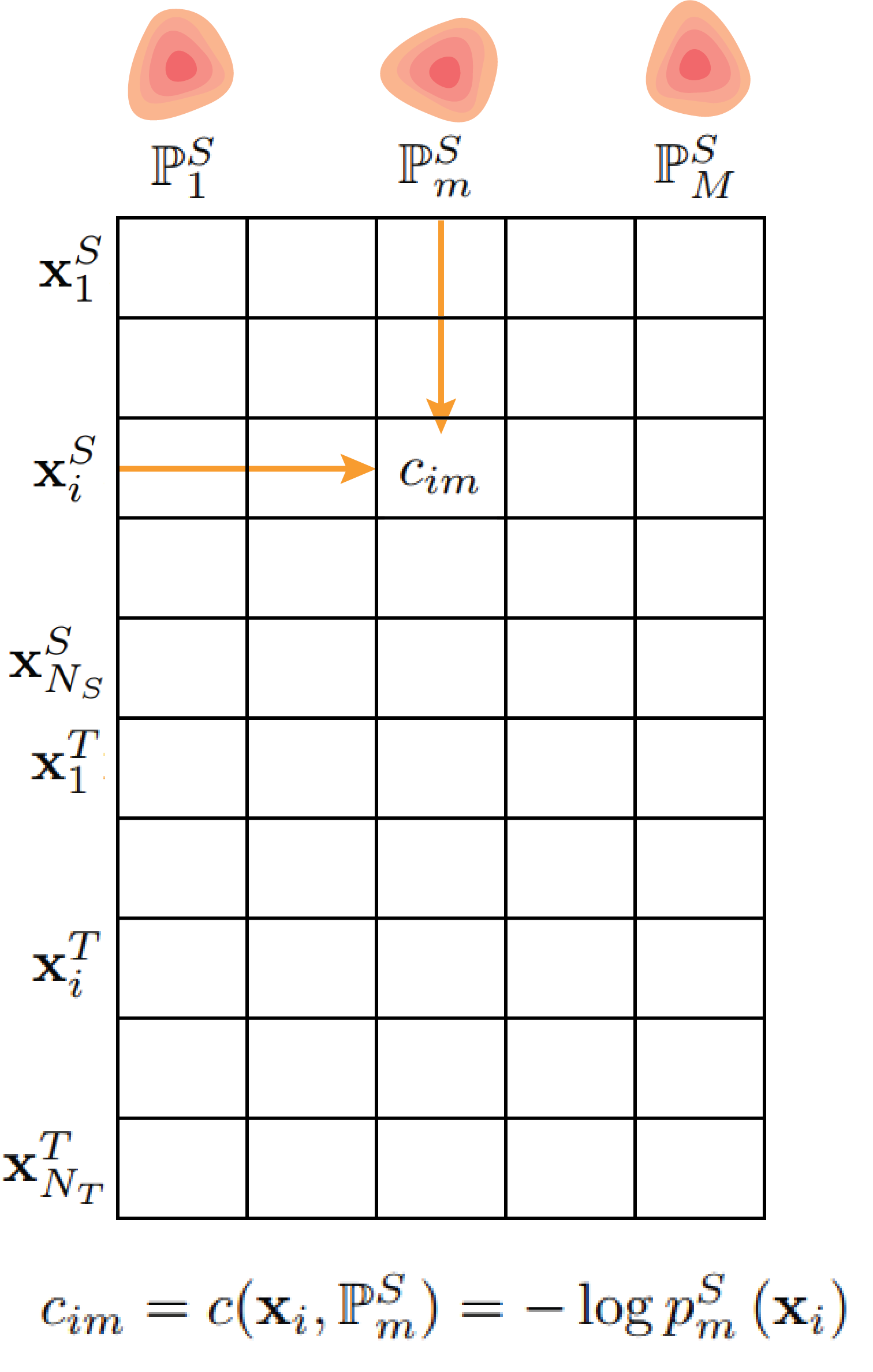}}\caption{The OT distance between two distributions: $\mathbb{P}$ and $\mathcal{P}^{S}$.
$\mathbb{P}$ consists of atoms representing the source and target
examples $\protect\bx_{i}$, while $\mathcal{P}^{S}$ consists of
atoms representing the source class-conditional distribution $\mathbb{P}_{m}^{S}$.\label{fig:matching}}
\end{figure}

\subsection{Our Proposed Class-aware OT}

In the given scenario with a total of $N$ samples, comprising $N_{S}$
samples from the source domain and $N_{T}$ samples from the target
domain, the source samples are denoted as $\mathbf{x}_{i}=\mathbf{x}_{i}^{S}$
for $1\leq i\leq N_{S}$, and the target samples are denoted as $\mathbf{x}_{i}=\mathbf{x}_{i-N_{S}}^{T}$
for $N_{S}+1\leq i\leq N$. To capture the distribution of both domains,
we define an empirical mixture distribution $\mathbb{P}$ as follows:

\[
\mathbb{P}=\frac{N_{S}}{N}\mathbb{P}^{S}+\frac{N_{T}}{N}\mathbb{P}^{T}=\frac{1}{N}\sum_{i=1}^{N}\delta_{\bx_{i}},
\]
where $\delta_{\mathbf{x}}$ represents the Dirac delta distribution
concentrated at the sample $\mathbf{x}$. Additionally, we introduce
$\mathcal{P}^{S}=\sum_{m=1}^{M}\pi_{m}\delta_{\mathbb{P}_{m}^{S}}$,
where the mixing proportion $\bpi\in\simplex_{M}$ (i.e., the $M-1$
dimensional simplex). In other words, $\mathcal{P}^{S}$ is a discrete
distribution of distributions, taking $\mathbb{P}_{m}^{S}$ with the
probability $\pi_{m}$.

We now examine an OT distance between two discrete distributions\footnote{We present the background of OT for two discrete distributions in
Appendix \ref{sec:Background}.}: $\mathbb{P}$ and $\mathcal{P}^{S}$. Our goal is to \emph{match
both source and target examples to the source class-conditional distributions,
}where a source example is explicitly guided to match the source class-conditional
distribution corresponding to its ground-truth label. In Section \ref{subsec:Amortization-Solution},
we provide further explanation on the motivation behind transporting
both source and target examples to $\mathcal{P}^{S}$.

Let us denote the cost $c\left(\bx_{i},\mathbb{P}_{m}^{S}\right)=-\log p_{m}^{S}\left(\bx_{i}\right)$
to match $\bx_{i}$ to $\mathbb{P}_{m}^{S}$ as $-\log p_{m}^{S}\left(\bx_{i}\right)$.
Specifically, if $\bx_{i}$ is more likely to be a sample from $\mathbb{P}_{m}^{S}$,
the log likelihood $\log p_{m}^{S}\left(\bx_{i}\right)$ is higher,
resulting in a smaller cost $c\left(\bx_{i},\mathbb{P}_{m}^{S}\right)$
(see Figure \ref{fig:matching}). We denote $A=\left[a_{im}\right]\in\mathbb{R}^{N\times M}$
as the transportation matrix, where $a_{im}$ represents the probability
of matching or transporting $\bx_{i}$ to $\mathbb{P}_{m}^{S}$. The
OT distance between $\mathbb{P}$ and $\mathcal{P}^{S}$ w.r.t. the
cost function $c$ and the mixing proportion $\bpi$ is defined as:

\begin{align}
\mathcal{W}_{c,\bpi}\left(\mathbb{P},\mathcal{P}^{S}\right) & =\min_{A}\Bigg\{\sum_{i=1}^{N}\sum_{m=1}^{M}a_{im}c\left(\bx_{i},\mathbb{P}_{m}^{S}\right)\nonumber \\
 & :\sum_{m=1}^{M}a_{im}=\frac{1}{N},\sum_{i=1}^{N}a_{im}=\pi_{m}\Bigg\}.\label{eq:ws_data}
\end{align}

Similar to other DA works \cite{pan2008transferlearning,TzengHDS15,long2017JAN},
we employ a feature extractor $G$ to map both source and target examples
to a latent space. We denote $\mathbb{Q}^{S},$ $\mathbb{Q}_{m}^{S},$
$\mathbb{Q}^{T},$ $\mathbb{Q},$ and $\mathcal{Q}^{S}$ as the corresponding
distributions over the latent space induced by $\mathbb{P}^{S},$
$\mathbb{P}_{m}^{S},$ $\mathbb{P}^{T},$ $\mathbb{P},$ and $\mathcal{P}^{S}$,
respectively, via the feature extractor $G$. The OT distance in Eq.
(\ref{eq:ws_data}) is rewritten as: 
\begin{align}
\mathcal{W}_{c,\bpi}\left(\mathbb{Q},\mathcal{Q}^{S}\right) & =\min_{A}\Bigg\{\sum_{i=1}^{N}\sum_{m=1}^{M}a_{im}c\left(G\left(\bx_{i}\right),\mathbb{Q}_{m}^{S}\right)\nonumber \\
 & :\sum_{m=1}^{M}a_{im}=\frac{1}{N},\sum_{i=1}^{N}a_{im}=\pi_{m}\Bigg\}.\label{eq:ws_latent}
\end{align}

\noindent To conduct domain adaptation, we aim to transport the target
examples $G\left(\bx_{i}\right)\,(N_{S}+1\leq i\leq N)$ to appropriate
class regions of the source domain. This can be achieved by solving
the following optimization problem: 
\begin{equation}
\min_{G,\bpi}\mathcal{W}_{c,\bpi}\left(\mathbb{Q},\mathcal{Q}^{S}\right).\label{eq:OT_Q}
\end{equation}

Given a data sample $G\left(\bx_{i}\right)$, let $\ba_{i}=\left[a_{im}\right]_{m}$
be its probability mass (i.e., the $i$-th row of the transportation
matrix). By setting the cost function as $c\left(G\left(\bx_{i}\right),\mathbb{Q}_{m}^{S}\right)=-\log p_{m}^{S}\left(\bx_{i}\right)$,
minimizing the OT distance in (\ref{eq:OT_Q}) encourages the following: 
\begin{enumerate}
\item The source sample $G\left(\bx_{i}\right)=G\left(\bx_{i}^{S}\right)\,$
for $1\leq i\leq N_{S}$ is matched exactly to $\mathbb{Q}_{y_{i}^{S}}^{S}$,
resulting in $\ba_{i}=\frac{\bone_{y_{i}^{S}}}{N}$, where $\mathbf{1}_{y}$
denotes the one-hot vector with the $y$-th element being one. 
\item The target sample $G\left(\bx_{i}\right)=G\left(\bx_{i-N_{S}}^{T}\right)\,$
for $N_{S}+1\leq i\leq N$ is encouraged to move to a $\mathbb{Q}_{k}^{S}$
distribution (where $1\leq k\leq M$) with a high likelihood. In this
case, $\mathbf{a}_{i}$ is inspired to be close to the corresponding
scaled one-hot vector $\frac{\mathbf{1}_{k}}{N}$. 
\end{enumerate}
Minimizing the OT distance through this optimization process encourages
the clustering of both source and target samples under their respective
source class distributions. In the subsequent sections, we will discuss
how to solve the optimization problem in (\ref{eq:OT_Q}) and the
definition of the cost function $c(G(\mathbf{x}_{i}),\mathbb{Q}_{m}^{S})$.

\subsection{Our Amortization Solution}

\label{subsec:Amortization-Solution}

The optimization problem in (\ref{eq:OT_Q}) can be solved analytically
using the Sinkhorn algorithm \cite{cuturi2013sinkhorn}. However,
this approach becomes computationally infeasible due to its high time
complexity of $\mathcal{O}(N^{2})$ for each iteration. To address
this computational challenge, one has explored a mini-batch version
of the Sinkhorn algorithm. Nevertheless, this approach may introduce
batch bias and yield inaccurate solutions \cite{nguyen2021transportation}.

In order to achieve an accurate distribution-matching solution and
enable batch training with a theoretical guarantee (cf. Theorem \ref{thm:amortize}),
we propose an amortization solution for minimizing the OT distance
in (\ref{eq:OT_Q}). Specifically, we employ a deep network called
the transportation network $\mathcal{T}\left(\cdot\right)$ with $M$
outputs, which allows us to amortize the transportation matrix $A$.

Given $\bx_{i}\,$, for $1\leq i\leq N$, we use the transportation
network $\mathcal{T}\left(G\left(\bx_{i}\right)\right))$ to predict
the probability $\ba_{i}=\left[a_{im}\right]_{m}$, where $a_{im}=\frac{1}{N}\mathcal{T}_{m}\left(G\left(\bx_{i}\right)\right)$.
This probability distribution specifies the likelihood of matching
or transporting $G\left(\bx_{i}\right)$ to $\mathbb{Q}_{m}^{S}$.
The constraint $\sum_{m=1}^{M}a_{im}=\frac{1}{N}$ is naturally satisfied.
Consequently, the optimization problem in (\ref{eq:OT_Q}) is rewritten
as: 
\begin{equation}
\min_{G,\mathcal{T}}\sum_{i=1}^{N}\sum_{m=1}^{M}\mathcal{T}_{m}\left(G\left(\bx_{i}\right)\right)c\left(G\left(\bx_{i}\right),\mathbb{Q}_{m}^{S}\right).\label{eq:amortized_1}
\end{equation}

The following theorem justifies solving the optimization problem in
(\ref{eq:amortized_1}) rather than directly solving the one in (\ref{eq:OT_Q}).
Specifically, we provide a theoretical proof demonstrating that when
the transportation network $\mathcal{T}$ is sufficiently expressive,
the optimal solution $\mathcal{T}^{*}$ obtained from (\ref{eq:amortized_1})
can effectively approximate the optimal transportation matrix $A^{*}$
derived from (\ref{eq:OT_Q}).
\begin{thm}
\label{thm:amortize}(Proof is presented in Appendix \ref{sec:All_proofs}).
Assuming that the transportation network $\mathcal{T}$ belongs to
a family of models with infinite capacity, which means it has the
ability to approximate any continuous function with arbitrary precision,
then the optimization problem in (\ref{eq:amortized_1}) is equivalent
to the optimization problem in (\ref{eq:OT_Q}). 
\end{thm}
The introduction of the network $\mathcal{T}$ also motivates us to
transport not only target examples but also source examples to their
respective source class-conditional distributions. This approach strengthens
the training of the network $\mathcal{T}$ on both the source and
target domains, allowing it to leverage its sufficient power and capacity
to effectively amortize the transportation matrix $A$.

We now present optimization problems related to transportation network
$\mathcal{T}$. For a source sample $G\left(\bx_{i}\right)=G\left(\bx_{i}^{S}\right)\,$,
where $1\leq i\leq N_{S}$, the prediction $\text{\ensuremath{\mathcal{T}\left(G\left(\bx_{i}\right)\right)}}$
should be the one-hot vector $\bone_{y_{i}^{S}}$. Therefore, we can
rewrite the optimization problem in (\ref{eq:amortized_1}) as follows:
\begin{align}
\min_{G,\mathcal{T}}\Bigg\{\mathcal{L}^{t}=\sum_{i=N_{S}+1}^{N}\sum_{m=1}^{M}\mathcal{T}_{m}\left(G\left(\bx_{i}\right)\right)c\left(G\left(\bx_{i}\right),\mathbb{Q}_{m}^{S}\right)\nonumber \\
+\sum_{i=1}^{N_{S}}CE\left(\bone_{y_{i}^{S}},\mathcal{T}\left(G\left(\bx_{i}\right)\right)\right)\Bigg\},\label{eq:amortized_2}
\end{align}
where $CE$ is the cross-entropy loss. For a target sample $G\left(\bx_{i}\right)=G\left(\bx_{i-N_{S}}^{T}\right)\,$,
where $N_{S}+1\leq i\leq N$, we propose a loss function inspired
by \cite{liang2020shot} to achieve the following objectives: 
\begin{enumerate}
\item Minimize the entropy of the prediction $\mathcal{T}\left(G\left(\bx_{i}\right)\right)$
to encourage the transportation network to make the clear decisions
of about where to move $G\left(\bx_{i}\right)$. 
\item Maximize the entropy of the average of \\
 $\sum_{i=N_{S}+1}^{N}\mathcal{T}\left(G\left(\bx_{i}\right)\right)$
for all target samples to encourage an equal movement to the class
regions. 
\end{enumerate}
To achieve these objectives, we minimize the following term: 
\begin{align}
\mathcal{L}^{ent} & =\frac{1}{N_{T}}\sum_{i=N_{S}+1}^{N}\left[\mathbb{H}\left(\mathcal{T}\left(G\left(\bx_{i}\right)\right)\right)\right]\nonumber \\
 & -\mathbb{H}\left(\frac{1}{N_{T}}\sum_{i=N_{S}+1}^{N}\mathcal{T}\left(G\left(\bx_{i}\right)\right)\right),
\end{align}
where $\mathbb{H}$ denotes the entropy function.

\subsection{Cost Function and Adversarial Training Framework}

In what follows, we explain how to define the cost function $c\left(G\left(\bx_{i}\right),\mathbb{Q}_{m}^{S}\right)=-\log p_{m}^{S}\left(\bx_{i}\right)$.
Drawing inspiration from the adversarial training framework \cite{goodfellow2014generative},
which has proven effective in unsupervised domain adaptation (UDA)
\cite{tzeng2017ADDA,long2018conditional,hoffman2018CyCADA}, we propose
the use of a multi-class discriminator $\mathcal{D}$. The purpose
of $\mathcal{D}$ is twofold: i) to provide a reliable measure for
defining the cost $c\left(G\left(\mathbf{x}{i}\right),\mathbb{Q}_{m}^{S}\right)$,
and ii) to enhance the alignment between $G\left(\bx_{i}\right)$
and the class-conditional distribution $\mathbb{Q}_{m}^{S}$ (i.e.,
the higher the alignment, the lower the cost) by encouraging target
samples to align with the entire source domain. To achieve these objectives,
we design $\mathcal{D}$ with $M+1$ outputs. For an example $\bx_{i}$,
$\mathcal{D}_{m}\left(G\left(\bx_{i}\right)\right)\,(1\leq m\leq M)$
represents the probability that $\bx_{i}$ belongs to the source domain
and has the label $m$, while $\mathcal{D}_{M+1}\left(G\left(\bx_{i}\right)\right)$
specifies the probability that $\bx_{i}$ belongs to the target domain.
This implies that $\sum_{m=1}^{M}\mathcal{D}_{m}\left(G\left(\bx_{i}\right)\right)=1-\mathcal{D}_{M+1}\left(G\left(\bx_{i}\right)\right)$
specifies the probability that $\bx_{i}$ belongs to the source domain.

The discriminator $\mathcal{D}$ is trained to distinguish the source
from target examples and predict the labels for source examples. The
loss to train the discriminator $\mathcal{D}$ is as follows: 
\begin{align}
\mathcal{L}^{\mathcal{D}} & =-\frac{1}{N_{T}}\sum_{i=N_{S}+1}^{N}\log\mathcal{D}_{M+1}\left(G\left(\bx_{i}\right)\right)\nonumber \\
 & \,\,\,\,\,\,-\frac{1}{N_{S}}\sum_{i=1}^{N_{S}}\log\left(1-\mathcal{D}_{M+1}\left(G\left(\bx_{i}\right)\right)\right)\nonumber \\
 & \,\,\,\,\,\,+\frac{1}{N_{S}}\sum_{i=1}^{N_{S}}CE\left(1_{y_{i}^{S}},\mathcal{D}_{1:M}\left(G\left(\bx_{i}\right)\right)\right),\label{eq:train_D}
\end{align}
where $\mathcal{D}_{1:M}\left(G\left(\bx_{i}\right)\right)=\left[\mathcal{D}_{m}\left(G\left(\bx_{i}\right)\right)\right]_{m=1}^{M}.$

The cost function in optimization problem (\ref{eq:amortized_2})
captures the extent to which $G\left(\bx_{i}\right)$ matches, and
we rely on the output of the discriminator $\mathcal{D}$ to define
$c\left(G\left(\bx_{i}\right),\mathbb{Q}_{m}^{S}\right)=-\log\mathcal{D}_{m}\left(G\left(\bx_{i}\right)\right)$
reasonably. Therefore, the loss function in (\ref{eq:amortized_2})
can be expressed as follows:

\begin{align}
\mathcal{L}^{t} & =-\sum_{i=N_{S}+1}^{N}\sum_{m=1}^{M}\mathcal{T}_{m}\left(G\left(\bx_{i}\right)\right)\log\mathcal{D}_{m}\left(G\left(\bx_{i}\right)\right)\nonumber \\
 & \,\,\,\,\:\,+\sum_{i=1}^{N_{S}}CE\left(\bone_{y_{y}^{S}},\mathcal{T}\left(G\left(\bx_{i}\right)\right)\right).
\end{align}

With more specific guidance from the discriminator $\mathcal{D}$,
the feature extractor $G$ is trained to push the target examples
$G\left(\bx_{i}\right),(N_{S}+1\leq i\leq N)$ to the \emph{entire
region} of source examples, and vice versa.

\textbf{For source examples.} Our objective is to move the source
examples $G\left(\bx_{i}\right),(1\leq i\leq N_{S})$ to the entire
region of the target examples characterized by high values of $\mathcal{D}_{M+1}$.
Therefore, we propose minimizing the following term:

\begin{equation}
\mathcal{L}^{G,S}=-\frac{1}{N_{S}}\sum_{i=1}^{N_{S}}\log\mathcal{D}_{M+1}\left(G\left(\bx_{i}\right)\right).\label{eq:LGS}
\end{equation}

\textbf{For target examples.} Our goal is to move the target examples
$G\left(\bx_{i}\right),(N_{S}+1\leq i\leq N)$ to the entire region
of the source examples characterized by low values of $\mathcal{D}_{M+1}$.
To achieve this, we minimize the following term:

\begin{equation}
\mathcal{L}^{G,T}=\frac{1}{N_{T}}\sum_{i=N_{S}+1}^{N}\log\mathcal{D}_{M+1}\left(G\left(\bx_{i}\right)\right).\label{eq:LGT}
\end{equation}

Furthermore, we utilize a source classifier $\mathcal{C}$ that is
trained on the source domain: 
\begin{equation}
\mathcal{L}^{C}=\frac{1}{N_{S}}\sum_{i=1}^{N_{S}}CE\left(\bone_{y_{i}^{S}},\mathcal{C}\left(G\left(\bx_{i}\right)\right)\right).
\end{equation}

Networks $\mathcal{C},G$, and $\mathcal{T}$ are trained by minimizing
the loss function: 
\begin{equation}
\mathcal{L}^{C}+\mathcal{L}^{G,S}+\mathcal{L}^{G,T}+\alpha\mathcal{L}^{t}+\beta\mathcal{L}^{ent},
\end{equation}
where $\alpha,\beta>0$ are trade-off parameters. We alternatively
update $\mathcal{D}$ by minimizing the loss function $\mathcal{L}^{\mathcal{D}}.$

According to \cite{farnia2018convex}, minimizing the Wasserstein
distance as in (\ref{eq:ws_latent}) encourages mode-covering behavior,
where the target examples tend to scatter over all modes of the source
data. On the other hand, minimizing the GAN loss in (\ref{eq:LGS},
\ref{eq:LGT}) or the Jensen-Shannon (JS) divergence encourages mode-seeking
behavior, where the target examples tend to cover specific modes of
the source data. Therefore, by minimizing $\mathcal{L}^{t}$ (a surrogate
of the Wasserstein distance) and $\mathcal{L}^{G,S}+\mathcal{L}^{G,T}$
(a surrogate of the JS divergence), we aim to encourage both mode-seeking
and mode-covering behaviors to achieve a more precise matching of
the source and target data.

\begin{figure*}[t]
\begin{centering}
\includegraphics[width=0.9\textwidth]{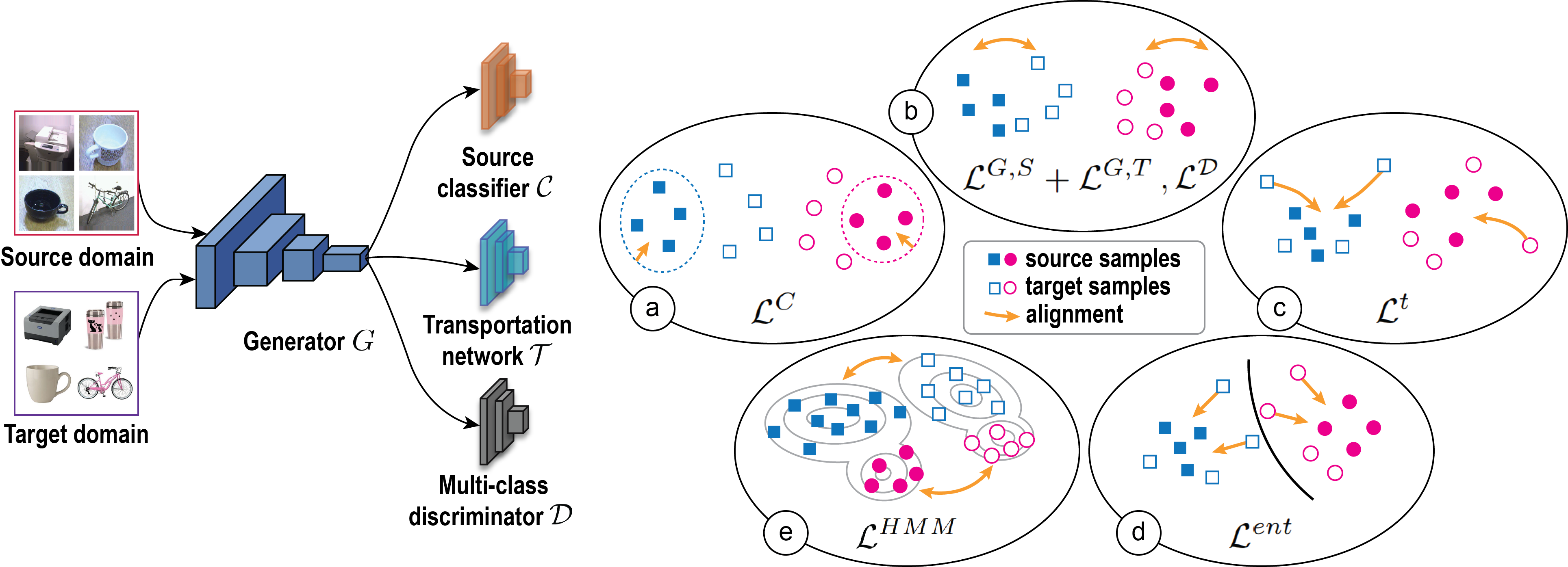}
\par\end{centering}
\vspace{2mm}

\caption{The framework of our proposed CLOTH consists of four components: a
weight-sharing generator $G$ for mapping the source and target data
into the latent space, a source classifier $\mathcal{C}$, a transportation
network $\mathcal{T}$, and a multi-class discriminator $\mathcal{D}$.
The model is trained by minimizing component losses $\mathcal{L}^{C},\mathcal{L}^{G,S}+\mathcal{L}^{G,T},\mathcal{L}^{\mathcal{D}},\mathcal{L}^{t}$
and $\mathcal{L}^{ent}$. (a) The classification loss $\mathcal{L}^{C}$
is minimized to accurately classify the source data with labels, resulting
in clear decision boundaries and well-clustered source samples. (b)
The generator $G$ and the multi-class discriminator $\mathcal{D}$
are trained adversarially, with updates alternated between $\mathcal{L}^{G,S}+\mathcal{L}^{G,T}$
and $\mathcal{L}^{\mathcal{D}}$. Unlike previous adversarial methods,
the source and target samples are mixed-up in a class-aware manner,
as further discussed in the ablation study. (c) The transportation
loss $\mathcal{L}^{t}$ is minimized to transport target samples to
the source class-conditional distribution, approximately minimizing
$\mathcal{W}_{c,\protect\bpi}\left(\mathbb{Q},\mathcal{Q}^{S}\right)$
in (\ref{eq:OT_Q}). (d) For target samples lying close to the decision
boundary, $\mathcal{T}$ is strengthened to provide a confident transportation
probability by minimizing $\mathcal{L}^{ent}$. This ensures that
these target samples are equally aligned to all source class regions.
(e) Class-aware HMM is proposed to more accurately capture the complex
distributions of the source and target domains in the latent space,
and therefore the class-aware matching between the two domains is
further improved.\label{fig:The-architecture-of-dotca}}
\end{figure*}

\subsection{Higher-Order Moment Matching}

\input{hmm.tex}

\subsection{Training Procedure of CLOTH}

By incorporating the loss $\mathcal{L}^{HMM}$, the loss to update
$\mathcal{C},\mathcal{T},$ and $G$ becomes: 
\begin{equation}
\mathcal{L}^{C}+\mathcal{L}^{G,S}+\mathcal{L}^{G,T}+\alpha\mathcal{L}^{t}+\beta\mathcal{L}^{ent}+\gamma\mathcal{L}^{HMM},\label{eq:final_obj}
\end{equation}
where $\gamma>0$ is a trade-off parameter. Finally, we present the
training algorithm of our proposed CLOTH in Algorithm \ref{alg:algorithm},
while the overall architecture and the motivation behind each component
loss are depicted in Figure \ref{fig:The-architecture-of-dotca}.

\begin{algorithm}[h]
\begin{algorithmic}[1]

\REQUIRE Source $\mathbb{D}_{S}=\left\{ \left(\bx_{i}^{S},y_{i}^{S}\right)\right\} _{i=1}^{N_{S}}$,
target $\mathbb{D}_{T}=\left\{ \bx_{j}^{T}\right\} _{j=1}^{N_{T}}$.
The number of training iterations $n_{t}$, batch size $b$, $q$-order,
and trade-off parameters $\alpha,\beta,\gamma$.

\ENSURE The optimal $\mathcal{C}^{*}$, $\mathcal{T}^{*}$, $G^{*}$,
and $\mathcal{D}^{*}$.

\FOR{$k=1$ to $n_{t}$}

\STATE Sample minibatch of source $\left\{ \left(\mathbb{\bx}_{i}^{S},y_{i}^{S}\right)\right\} _{i=1}^{b}$
and target $\left\{ \bx_{j}^{T}\right\} _{j=1}^{b}$.

\STATE Update $\mathcal{D}$ according to (\ref{eq:train_D}).

\STATE Sample minibatch of source $\left\{ \left(\mathbb{\bx}_{i}^{S},y_{i}^{S}\right)\right\} _{i=1}^{b}$
and target $\left\{ \bx_{j}^{T}\right\} _{j=1}^{b}$.

\STATE Update $\mathcal{C},\mathcal{T},$ and $G$ according to (\ref{eq:final_obj}).

\ENDFOR

\end{algorithmic} \caption{Pseudocode for training our proposed CLOTH.\label{alg:algorithm}}
\end{algorithm}

%% file: hmm.tex
To further enhance the matching extent of a target sample to a proper
source class distribution $\mathbb{Q}_{m}^{S}$, we leverage the higher-order
moments (HM) method which has been successfully employed for cross-domain
matching in neural style transfer \cite{gatys2016image,li2017demystifying},
knowledge distillation \cite{yim2017agift}, and domain adaptation
\cite{chen2020homm}. By considering higher-order statistics (greater
than second-order), we can achieve fine-grained domain alignment,
enabling the approximation of complex distributions in the latent
space of both the source and target domains. This fine-grained alignment
facilitates a more accurate matching process. In the upcoming sections,
we introduce an efficient solution for measuring higher-order moment
matching between the distributions of the source and target domains,
leading to our proposed class-aware higher-order moment matching method.

\subsection{Higher-Order Moment with Lower Complexity}

Given a vector $\bz=G\left(\bx\right)\in\mathbb{R}^{p}$, which represents
a data sample in an adapted layer. With $q\leq p$, we define a $q$-order
moment of $\bz$ is a $q$-dimensional tensor $A_{q}\left(\bz\right)\in\mathbb{R}^{p^{q}}$,
where: 
\begin{equation}
A_{q}\left(\bz\right)\left[i_{1},...,i_{q}\right]=\prod_{j=1}^{q}z_{i_{j}}
\end{equation}
for any index $i_{j=1,2,...,q}\in\left\{ 1,2,...,p\right\} .$ For
example, with $q=1$, $A_{1}\left(\bz\right)\in\mathbb{R}^{p},A_{1}\left(\bz\right)\left[i_{1}\right]=z_{i_{1}}$
and $q=2$, $A_{2}\left(\bz\right)\in\mathbb{R}^{p^{2}},A_{2}\left(\bz\right)\left[i_{1},i_{2}\right]=z_{i_{1}}z_{i_{2}}$.

Given two distributions $\mathbb{U}$ and $\mathbb{V}$, we consider
an HM distance between these two distributions as:

\begin{align}
HM\left(\mathbb{U},\mathbb{V}\right) & =\norm{\mathbb{E}_{\mathbb{U}}\left[A_{q}\left(\bz\right)\right]-\mathbb{E}_{\mathbb{V}}\left[A_{q}\left(\bz\right)\right]}_{F}^{2},\label{eq:HM_PQ_tensor}
\end{align}

where $\norm ._{F}$ represents a Frobenius norm. Unfortunately, calculating
this HM distance is infeasible in practice due to the space complexity
reaching $\mathcal{O}\left(p^{q}\right)$, especially as the order
$q$ increases. To mitigate this complexity, \cite{chen2020homm}
proposed a random sampling matching strategy to perform arbitrary-order
moment matching between the distributions of the source and target
domains in the output layer. Specifically, the authors randomly selected
$n$ values from $A_{q}\left(\bz\right)$ and only matched these $n$
values in the source and target domains. This approach reduces the
space complexity to $\mathcal{O}\left(n\right)$. However, there are
limitations in this method: (i) randomly sampling can lead to the
loss of important higher-order statistics that potentially characterize
the distribution; and (ii) the approach achieves competitive results
only when $n\geq1000$. To address these concerns, we propose a simple
and economical solution for more accurate and efficient computation
of the HM distance to facilitate higher-order moment matching. We
define $\phi_{q}\left(\bz\right)$ as the vector obtained by flattening
the $q$-dimensional tensor $A_{q}\left(\bz\right)$, which means
that:

\begin{equation}
\phi_{q}\left(\bz\right)=\left[\prod_{j=1}^{q}z_{i_{j}}\right]_{1\leq i_{1},...,i_{q}\leq p}.
\end{equation}
The HM distance in Eq. (\ref{eq:HM_PQ_tensor}) is rewritten as:\vspace{-1mm}
 
\begin{align}
HM\left(\mathbb{U},\mathbb{V}\right) & =\norm{\mathbb{E}_{\mathbb{U}}\left[\phi_{q}\left(\bz\right)\right]-\mathbb{E}_{\mathbb{V}}\left[\phi_{q}\left(\bz\right)\right]}_{2}^{2},\label{eq:HM_PQ_flatten}
\end{align}
where $\norm ._{2}$ represents the L2 norm. We now devise an economical
way to compute $HM\left(\mathbb{U},\mathbb{V}\right)$. We begin with
a well-known equality: 
\begin{equation}
\left(a_{1}+...+a_{p}\right)^{q}=\sum_{1\leq i_{1},...,i_{q}\leq p}\prod_{j=1}^{q}a_{i_{j}}.\label{eq:sum_q}
\end{equation}
Given $\bz$ and $\bz'$, the dot product $<\phi_{q}\left(\bz\right),\phi_{q}\left(\bz'\right)>$
can be further derived as 
\begin{equation}
\sum_{1\leq i_{1},...,i_{q}\leq p}\prod_{j=1}^{q}\left(\bz_{i_{j}}\bz_{i_{j}}^{'}\right)=\left(\sum_{i=1}^{p}\bz_{i}\bz'_{i}\right)^{q}=\left\langle \bz,\bz'\right\rangle ^{q},
\end{equation}
where we apply Eq. (\ref{eq:sum_q}) for $a_{i}=\bz_{i}\bz'_{i}$.
The HM distance $HM\left(\mathbb{U},\mathbb{V}\right)$ in Eq. (\ref{eq:HM_PQ_flatten})
can be further derived as

$\mathbb{E}_{\bz,\bz'\sim\mathbb{U}}\left[<\bz,\bz'>^{q}\right]+\mathbb{E}_{\bz,\bz'\sim\mathbb{V}}\left[<\bz,\bz'>^{q}\right]$\vspace{-5mm}

\begin{equation}
\,\,\,\,\,\,\,\,\,\:\,\,\,\,\,\,\,\,\,\,\,\,\,\,\,\,-2\mathbb{E}_{\bz\sim\mathbb{U},\bz'\sim\mathbb{V}}\left[<\bz,\bz'>^{q}\right].\label{eq:HM_PQ_kernel-1}
\end{equation}

It is important to note that the calculation of the HM distance in
Eq. (\ref{eq:HM_PQ_tensor}) becomes feasible, with the space complexity
reduced from $\mathcal{O}\left(p^{q}\right)$ to $\mathcal{O}\left(p\right)$.
In practice, $\mathbb{U}$ and $\mathbb{V}$ are empirical distributions,
allowing for a convenient approximation of $HM\left(\mathbb{U},\mathbb{V}\right)$
in Eq. (\ref{eq:HM_PQ_kernel-1}) based on mini-batches.

\subsection{Class-aware Higher-Order Moment Matching}

\label{sec:CaHoMM}

We leverage the transportation network $\mathcal{T}$ to propose a
more elegant loss as:

\noindent \resizebox{1.0\columnwidth}{!}{$\mathcal{L}^{HMM}=\frac{1}{M}\sum_{m=1}^{M}\norm{\mathbb{E}_{\mathbb{Q}_{m}^{S}}\left[\phi_{q}\left(\bz\right)\right]-\mathbb{E}_{\mathbb{Q}^{T}}\left[\mathcal{T}_{m}\left(\bz\right)\phi_{q}\left(\bz\right)\right]}_{2}^{2}$}

\noindent $=\frac{1}{M}\sum_{m=1}^{M}$

\noindent 
\begin{equation}
\norm{\mathbb{E}_{\mathbb{P}_{m}^{S}}\left[\phi_{q}\left(G\left(\bx\right)\right)\right]-\mathbb{E}_{\mathbb{P}^{T}}\left[\mathcal{T}_{m}\left(G\left(\bx\right)\right)\phi_{q}\left(G\left(\bx\right)\right)\right]}_{2}^{2}.\label{eq:class_aware}
\end{equation}

The objective function in (\ref{eq:class_aware}) aims to match a
class in the source domain and the corresponding class in the target
domain with the guidance from the transformation network. The transportation
network $\mathcal{T}_{m}\left(G\left(\bx\right)\right)$ provides
a value close to $1$ if the target example $\bx$ appears to belong
to class $m$. Expanding the above expression, we arrive at: 
\begin{multline}
\mathcal{L}^{HMM}=\frac{1}{M}\sum_{m=1}^{M}\biggl(\mathbb{E}_{\bx,\bx'\sim\mathbb{P}_{m}^{S}}\left[<G\left(\bx\right),G\left(\bx'\right)>^{q}\right]\\
+\mathbb{E}_{\bx,\bx'\sim\mathbb{P}^{T}}\left[\mathcal{T}_{m}\left(G\left(\bx\right)\right)\mathcal{T}_{m}\left(G\left(\bx'\right)\right)<G\left(\bx\right),G\left(\bx'\right)>^{q}\right]\\
-2\mathbb{E}_{\bx\sim\mathbb{P}_{m}^{S},\bx'\sim\mathbb{P}^{T}}\left[\mathcal{T}_{m}\left(G\left(\bx'\right)\right)<G\left(\bx\right),G\left(\bx'\right)>^{q}\right]\biggr).\label{eq:L_HMM}
\end{multline}

%% file: experiment.tex
In this section, we perform experiments on four benchmark domain adaptation
datasets: \emph{Digits}, \emph{Office-31}, \emph{Office-Home}, and
\emph{ImageCLEF-DA}. We compare our CLOTH with a variety of baselines
including the standard baseline ResNet-50 \cite{he2016resnet} and
existing works including DAN \cite{long2015}, DANN \cite{Ganin2015},
RTN \cite{long2016rtn}, iCAN \cite{zhang2018ican}, CDAN-E \cite{long2018cdan},
CDAN-BSP \cite{pmlr-v97-chen19i}, CDAN-TransNorm (CDAN-TN) \cite{Wang2019CDAN},
CADA-P \cite{kurmi2019attending}, SymNets \cite{zhang2019symnets},
especially class-aware DA and OT-based methods, namely RADA \cite{Wang2019classaware},
CAN \cite{Kang2019classaware}, DeepJDOT \cite{damodaran2018deepjdot},
ETD \cite{MengxueETD2020}, RWOT \cite{xu2020reliable}, and LAMDA
\cite{le2021lamda}. Furthermore, we provide the information of the
network architecture and additional ablation studies in Appendix \ref{sec:Net_arch}
and \ref{sec:Add_anal}, respectively.

\subsection{Datasets}

\emph{Digits} is a standard DA dataset that consists of four widely
used benchmarks: \textbf{MNIST} \cite{lecun1998lenet}, \textbf{USPS}
\cite{Hull_usps_1994}, Street View House Numbers (\textbf{SVHN})
\cite{netzer2011svhn}, and synthetic digits dataset (\textbf{SYN})
\cite{ganin2016domainadversarial}. In order to evaluate and compare
the performance of our proposed methods with the baselines, we focus
on three typical transfer tasks: \textbf{SVHN}$\rightarrow$\textbf{MNIST},
\textbf{SYN}$\rightarrow$\textbf{MNIST}, and \textbf{USPS}$\rightarrow$\textbf{MNIST}.

\emph{Office-31} \cite{Saenkooffice312010} is is a widely recognized
public dataset extensively used for UDA. It comprises three domains:
Amazon (\textbf{A}), which contains product images obtained from amazon.com;
Webcam (\textbf{W}), consisting of low-resolution images captured
by a webcam; and Dslr (\textbf{D}), comprising high-resolution images
taken by a digital SLR camera. The dataset encompasses 31 common classes
and a total of 4,110 images. Specifically, the Amazon domain contains
2,817 images, the Webcam domain contains 795 images, and the Dslr
domain contains 498 images.

\emph{Office-Home} \cite{venkateswara2017deephasing} is a challenging
dataset specifically designed for unsupervised domain adaptation (UDA).
It comprises images from four distinct domains: Artistic (\textbf{Ar}),
Clip Art (\textbf{Cl}), Product (\textbf{Pr}), and Real-world images
(\textbf{Re}). The dataset encompasses approximately 15,588 images
in total, covering 65 object categories within office and home scenes.

The final dataset used in our experiments is the \emph{ImageCLEF-DA}
dataset \cite{CaputoImageCLEF2014}. This dataset comprises three
domains: Caltech-256 (\textbf{C}), ImageNet ILSVRC 2012 (\textbf{I}),
and Pascal VOC 2012 (\textbf{P}). Each domain consists of 12 classes,
with 50 images per class.

\subsection{Implementation Details}

In our experiments on the \emph{Office-31}, \emph{Office-Home} and
\emph{ImageCLEF-DA} datasets, we use the extracted features from ResNet-50
\cite{he2016resnet}. For \emph{Digits}, we use the modified version
of LeNet \cite{lecun1998lenet} as used in previous work \cite{chen2020homm}.
Our proposed methods are trained using the Adam \cite{KingmaB14}
optimizer with Polyak averaging \cite{polyak1992acce} in Tensorflow
\cite{abadi2016tensorflow}. In the ablation study, we select the
trade-off hyper-parameters $\alpha$ and $\beta$ from the set $\left\{ 10^{-2},10^{-1},10^{0}\right\} $
for \emph{Office-31}, \emph{ImageCLEF-DA} and \emph{Office-Home},
and from $\left\{ 10^{-3},10^{-2},10^{-1}\right\} $ for the \emph{Digits}.
The hyperparameter $\gamma$ is selected from $\left\{ 10^{-3},10^{-2},10^{-1}\right\} $.
We set the value of the $q$-order moment to 3, the learning rate
to $10^{-4}$, and the mini-batch size to 128. After the training
process, we choose the best parameter set based on the validation
accuracy on the source domain. Finally, we run each transfer task
5 times and report the average accuracy.

\subsection{Result and Discussion}

\begin{table}
\centering{}\caption{Classification accuracy (\%) on Digits for UDA (LeNet).\label{tab:The-test-accuracy_digits}}
\resizebox{1.00\columnwidth}{!}{%%
\begin{tabular}{ccccc}
\hline 
Method  & SVHN$\rightarrow$MNIST  & SYN$\rightarrow$MNIST  & USPS$\rightarrow$MNIST  & Avg\tabularnewline
\hline 
LeNet \cite{lecun1998lenet}  & 67.3  & 66.4  & 89.7  & 74.5\tabularnewline
DDC \cite{TzengHZSD14}  & 71.9  & 75.8  & 89.9  & 79.2\tabularnewline
DAN \cite{long2015}  & 79.5  & 89.8  & 75.2  & 81.5\tabularnewline
DANN \cite{ganin2016domainadversarial}  & 70.6  & 76.6  & 90.2  & 79.1\tabularnewline
CMD \cite{zellinger2019central}  & 86.5  & 86.3  & 96.1  & 89.6\tabularnewline
ADDA \cite{tzeng2017ADDA}  & 72.3  & 92.1  & 96.3  & 86.9\tabularnewline
CORAL \cite{SunDCORAL2016}  & 89.5  & 96.5  & 96.5  & 94.2\tabularnewline
CyCADA \cite{HoffmanCyCADA2017}  & 92.8  & 97.4  & 97.5  & 95.9\tabularnewline
JDDA \cite{ChenJDAN2018}  & 94.2  & 96.7  & 97.7  & 96.2\tabularnewline
HoMM \cite{chen2020homm}  & 99.0  & 99.1  & 99.2  & 99.1\tabularnewline
\hline 
\textbf{CLOTH}  & \textbf{99.1}  & \textbf{99.5}  & \textbf{99.3}  & \textbf{99.3}\tabularnewline
\hline 
\end{tabular}} 
\end{table}

\begin{table}[t]
\centering{}\caption{Classification accuracy (\%) on Office-31 for UDA (ResNet-50).\label{tab:The-test-accuracy-office-31}}
\resizebox{1.0\columnwidth}{!}{%%
\begin{tabular}{cccccccc}
\hline 
Method  & A$\rightarrow$W  & A$\rightarrow$D  & D$\rightarrow$W  & W$\rightarrow$D  & D$\rightarrow$A  & W$\rightarrow$A  & Avg\tabularnewline
\hline 
ResNet-50 \cite{he2016resnet}  & 68.4  & 68.9  & 96.7  & 99.3  & 62.5  & 60.7  & 76.1\tabularnewline
DAN \cite{long2015}  & 80.5  & 78.6  & 97.1  & 99.6  & 63.6  & 62.8  & 80.4\tabularnewline
DANN \cite{Ganin2015}  & 82.0  & 79.7  & 96.9  & 99.1  & 68.2  & 67.4  & 82.2\tabularnewline
iCAN \cite{ZhangiCAN2018}  & 92.5  & 90.1  & 98.8  & \textbf{100.0}  & 72.1  & 69.9  & 87.2\tabularnewline
ADDA \cite{tzeng2017ADDA}  & 75.5  & 88.2  & 96.5  & 89.1  & 75.1  & 92.0  & 86.0\tabularnewline
CDAN \cite{long2018conditional}  & 94.1  & 92.9  & 98.6  & \textbf{100.0}  & 71.0  & 69.3  & 87.7\tabularnewline
SHOT \cite{liang2020shot}  & 90.1  & 94.0  & 98.4  & 99.0  & 74.7  & 74.3  & 88.6\tabularnewline
DeepJDOT \cite{damodaran2018deepjdot}  & 88.9  & 88.2  & 98.5  & 99.6  & 72.1  & 70.1  & 86.2\tabularnewline
ETD \cite{MengxueETD2020}  & 92.1  & 88.0  & \textbf{100.0}  & \textbf{100.0}  & 71.0  & 69.3  & 86.2\tabularnewline
RWOT \cite{xu2020reliable}  & 95.1  & 94.5  & 99.5  & \textbf{100.0}  & 77.5  & 77.9  & 90.8\tabularnewline
LAMDA \cite{le2021lamda}  & 95.2  & 96.0  & 98.5  & \textbf{100.0}  & 87.3  & 84.4  & 93.0\tabularnewline
HoMM \cite{chen2020homm}  & 91.7  & 89.1  & 98.8  & \textbf{100.0}  & 71.2  & 70.6  & 86.9\tabularnewline
RADA \cite{Wang2019classaware}  & 91.5  & 90.7  & 98.9  & \textbf{100.0}  & 71.5  & 71.3  & 87.3\tabularnewline
CAN \cite{Kang2019classaware}  & 94.5  & 95.0  & 99.1  & 99.8  & 78.0  & 77.0  & 90.6\tabularnewline
\hline 
\textbf{CLOTH}  & \textbf{96.9}  & \textbf{99.0}  & 98.1  & \textbf{100.0}  & \textbf{87.6}  & \textbf{86.5}  & \textbf{94.7}\tabularnewline
\hline 
\end{tabular}} 
\end{table}

\begin{table*}[t]
\centering{}\caption{Classification accuracy (\%) on Office-Home for UDA (ResNet-50).\label{tab:The-test-accuracy-officehome}}
\resizebox{1.00\textwidth}{!}{%%
\begin{tabular}{cccccccccccccc}
\hline 
Method  & Ar$\rightarrow$Cl  & Ar$\rightarrow$Pr  & Ar$\rightarrow$Re  & Cl$\rightarrow$Ar  & Cl$\rightarrow$Pr  & Cl$\rightarrow$Re  & Pr$\rightarrow$Ar  & Pr$\rightarrow$Cl  & Pr$\rightarrow$Re  & Re$\rightarrow$Ar  & Re$\rightarrow$Cl  & Re$\rightarrow$Pr  & Avg\tabularnewline
\hline 
ResNet-50 \cite{he2016resnet}  & 34.9  & 50.0  & 58.0  & 37.4  & 41.9  & 46.2  & 38.5  & 31.2  & 60.4  & 53.9  & 41.2  & 59.9  & 46.1\tabularnewline
DANN \cite{Ganin2015}  & 45.6  & 59.3  & 70.1  & 47.0  & 58.5  & 60.9  & 46.1  & 43.7  & 68.5  & 63.2  & 51.8  & 76.8  & 57.6\tabularnewline
DAN \cite{long2015}  & 43.6  & 57.0  & 67.9  & 45.8  & 56.5  & 60.4  & 44.0  & 43.6  & 67.7  & 63.1  & 51.5  & 74.3  & 56.3\tabularnewline
CDAN \cite{long2018conditional}  & 50.7  & 70.6  & 76.0  & 57.6  & 70.0  & 70.0  & 57.4  & 50.9  & 77.3  & 70.9  & 56.7  & 81.6  & 65.8\tabularnewline
SHOT \cite{liang2020shot}  & 57.1  & 78.1  & 81.5  & \textbf{68.0}  & 78.2  & 78.1  & \textbf{67.4}  & 54.9  & 82.2  & 73.3  & 58.8  & 84.3  & 71.8\tabularnewline
SPL \cite{WangSPL2019}  & 54.5  & 77.8  & 81.9  & 65.2  & 78.0  & 81.1  & 66.0  & 53.1  & 82.8  & 69.9  & 55.3  & \textbf{86.0}  & 71.0\tabularnewline
CADA-P \cite{KurmiCADA2019}  & 56.9  & 76.4  & 80.7  & 61.3  & 75.2  & 75.2  & 63.2  & 54.5  & 80.7  & \textbf{73.9}  & \textbf{61.5}  & 84.1  & 70.2\tabularnewline
RFA \cite{awais2021RFA}  & 55.7  & 77.1  & 80.7  & 64.4  & 74.8  & 75.5  & 64.0  & 53.1  & 80.6  & 71.8  & 58.4  & 84.3  & 70.0\tabularnewline
DeepJDOT \cite{damodaran2018deepjdot}  & 48.2  & 69.2  & 74.5  & 58.5  & 69.1  & 71.1  & 56.3  & 46.0  & 76.5  & 68.0  & 52.7  & 80.9  & 64.3\tabularnewline
ETD \cite{MengxueETD2020}  & 51.3  & 71.9  & \textbf{85.7}  & 57.6  & 69.2  & 73.7  & 57.8  & 51.2  & 79.3  & 70.2  & 57.5  & 82.1  & 67.3\tabularnewline
RWOT \cite{xu2020reliable}  & 55.2  & 72.5  & 78.0  & 63.5  & 72.5  & 75.1  & 60.2  & 48.5  & 78.9  & 69.8  & 54.8  & 82.5  & 67.6\tabularnewline
\hline 
\textbf{CLOTH}  & \textbf{57.3}  & \textbf{78.8}  & 82.8  & 67.4  & \textbf{80.4}  & \textbf{81.3}  & 66.5  & \textbf{56.6}  & \textbf{83.1}  & 71.7  & 59.5  & 84.0  & \textbf{72.5}\tabularnewline
\hline 
\end{tabular}} 
\end{table*}

\begin{table}
\centering{}\caption{Classification accuracy (\%) on ImageCLEF-DA for UDA (ResNet-50).\label{tab:The-test-accuracy-clef-da}}
\resizebox{1.0\columnwidth}{!}{%%
\begin{tabular}{cccccccc}
\hline 
Method  & I$\rightarrow$P  & P$\rightarrow$I  & I$\rightarrow$C  & C$\rightarrow$I  & C$\rightarrow$P  & P$\rightarrow$C  & Avg\tabularnewline
\hline 
ResNet-50 \cite{he2016resnet}  & 74.8  & 83.9  & 91.5  & 78.0  & 65.5  & 91.2  & 80.7\tabularnewline
RTN \cite{long2016rtn}  & 75.6  & 86.8  & 95.3  & 86.9  & 72.7  & 92.2  & 84.9\tabularnewline
ADDA \cite{tzeng2017ADDA}  & 75.5  & 88.2  & 96.5  & 89.1  & 75.1  & 92.0  & 86.0\tabularnewline
iCAN \cite{ZhangiCAN2018}  & 79.5  & 89.7  & 94.7  & 89.9  & 78.5  & 92.0  & 87.4\tabularnewline
CDAN-E \cite{long2018conditional}  & 77.7  & 90.7  & 97.7  & 91.3  & 74.2  & 94.3  & 87.7\tabularnewline
CDAN-TN \cite{Wang2019CDAN}  & 78.3  & 90.8  & 96.7  & 92.3  & 78.0  & 94.8  & 88.5\tabularnewline
SymNets \cite{zhang2019symnets}  & 80.2  & 93.6  & 97.0  & 93.4  & 78.7  & 96.4  & 89.9\tabularnewline
MEDA \cite{WangMEDA2018}  & 79.7  & 92.5  & 95.7  & 92.2  & 78.5  & 95.5  & 89.0\tabularnewline
SPL \cite{WangSPL2019}  & 78.3  & 94.5  & 96.7  & 95.7  & 80.5  & 96.3  & 90.3\tabularnewline
CADA-P \cite{KurmiCADA2019}  & 78.0  & 90.5  & 96.7  & 92.0  & 77.2  & 95.5  & 88.3\tabularnewline
A\textsuperscript{2}LP \cite{zhang2020label}  & 79.3  & 91.8  & 96.3  & 91.7  & 78.1  & 96.0  & 88.9\tabularnewline
DeepJDOT \cite{damodaran2018deepjdot}  & 77.7  & 90.6  & 95.1  & 88.5  & 75.3  & 94.3  & 86.9\tabularnewline
ETD \cite{MengxueETD2020}  & 81.0  & 91.7  & 97.9  & 93.3  & 79.5  & 95.0  & 89.7\tabularnewline
RWOT \cite{xu2020reliable}  & 81.5  & 93.1  & \textbf{98.0}  & 92.8  & 79.3  & \textbf{96.8}  & 90.3\tabularnewline
\hline 
\textbf{CLOTH}  & \textbf{83.2}  & \textbf{95.0}  & 97.5  & \textbf{95.8}  & \textbf{80.7}  & 96.7  & \textbf{91.5}\tabularnewline
\hline 
\end{tabular}} 
\end{table}

\begin{table}[h]
\centering{}\caption{Accuracy (\%) of ablation study on Office-31 and ImageCLEF-DA.\label{tab:effect_loss}}
\resizebox{1.0\columnwidth}{!}{\centering\setlength{\tabcolsep}{2pt}%{\small{}}%
\begin{tabular}{cccccccc}
\hline 
Method  & $\mathcal{L}^{C}$  & $\mathcal{L}^{G,S}+\mathcal{L}^{G,T}$  & $\mathcal{L}^{t}$  & $\mathcal{L}^{ent}$  & $\mathcal{L}^{HMM}$  & Office-31  & ImageCLEF-DA\tabularnewline
\hline 
1  & {\small{}{}\checkmark}  &  &  &  &  & 76.1  & 80.7\tabularnewline
2  & {\small{}{}\checkmark}  & {\small{}{}\checkmark}  &  &  &  & 87.7  & 86.9\tabularnewline
3  & {\small{}{}\checkmark}  & {\small{}{}\checkmark}  & {\small{}{}\checkmark}  &  &  & 92.5  & 88.6\tabularnewline
4  & {\small{}{}\checkmark}  & {\small{}{}\checkmark}  & {\small{}{}\checkmark}  & {\small{}{}\checkmark}  &  & 92.8  & 89.4\tabularnewline
5  & {\small{}{}\checkmark}  & {\small{}{}\checkmark}  & {\small{}{}\checkmark}  &  & {\small{}{}\checkmark}  & 92.9  & 89.2\tabularnewline
6  & {\small{}{}\checkmark}  & {\small{}{}\checkmark}  &  & {\small{}{}\checkmark}  & {\small{}{}\checkmark}  & 91.3  & 88.6\tabularnewline
7  & {\small{}{}\checkmark}  & {\small{}{}\checkmark}  & {\small{}{}\checkmark}  & {\small{}{}\checkmark}  & {\small{}{}\checkmark}  & \textbf{94.7}  & \textbf{91.5}\tabularnewline
\hline 
\end{tabular}}\vspace{-1mm}
 
\end{table}

The experimental results in Table \ref{tab:The-test-accuracy_digits}
demonstrate that our CLOTH achieves superior performance compared
to other state-of-the-art baselines on all transfer tasks in the \emph{Digits}
dataset. This includes tasks involving grayscale digits (e.g., \textbf{USPS}$\rightarrow$\textbf{MNIST})
as well as tasks where color digits are transferred to grayscale hand-written
digits (e.g., \textbf{SVHN}$\rightarrow$\textbf{MNIST}, \textbf{SYN}$\rightarrow$\textbf{MNIST}).

The results for the \emph{Office-31} dataset are presented in Table
\ref{tab:The-test-accuracy-office-31}. Our CLOTH method shows significant
improvements over the comparison methods in most of the transfer tasks
(\textbf{A}$\rightarrow$\textbf{D}, \textbf{W}$\rightarrow$\textbf{D},
\textbf{D}$\rightarrow$\textbf{A}, and \textbf{W}$\rightarrow$\textbf{A}).
Additionally, our CLOTH achieves the highest average accuracy of 94.7\%,
which is a significant improvement of 1.7\% compared to the runner-up
baseline (LAMDA).

Table \ref{tab:The-test-accuracy-officehome} presents the results
of our proposed CLOTH on the challenged \emph{Office-Home} dataset.
Our CLOTH method outperforms the baselines on 6 out of 12 representative
transfer tasks (\textbf{Ar}$\rightarrow$\textbf{Pr}, \textbf{Cl}$\rightarrow$\textbf{Pr},
\textbf{Cl}$\rightarrow$\textbf{Re}, \textbf{Pr}$\rightarrow$\textbf{Cl},
and \textbf{Pr}$\rightarrow$\textbf{Re}). In terms of average accuracy,
our CLOTH achieves the highest value of 72.5\%, followed by the SHOT
method with 71.8\%.

On the \emph{ImageCLEF-DA} dataset, our proposed CLOTH demonstrates
significant improvements over the baselines on four out of six transfer
tasks (\textbf{I}$\rightarrow$\textbf{P}, \textbf{P}$\rightarrow$\textbf{I},
\textbf{C}$\rightarrow$\textbf{I}, and \textbf{C}$\rightarrow$\textbf{P}),
as shown in Table \ref{tab:The-test-accuracy-clef-da}. Additionally,
CLOTH achieves the highest average accuracy of $91.5\%$, outperforming
other baselines, with RWOT being the closest at $90.3\%$.

\subsection{Analysis}

\subsubsection{Effect of Losses}

We conducted experiments to investigate the effectiveness of the component
losses $\mathcal{L}^{C}$, $\mathcal{L}^{G,S}+\mathcal{L}^{G,T}$,
$\mathcal{L}^{t}$, $\mathcal{L}^{ent}$, and $\mathcal{L}^{HMM}$
in (\ref{eq:final_obj}). The results are summarized in Table \ref{tab:effect_loss}.
In the source-only setting, only $\mathcal{L}^{C}$ is utilized (Method
1). The performance is significantly improved by 11.6\% on \emph{Office-31}
and 6.2\% on \emph{ImageCLEF-DA} when adversarial training with $\mathcal{L}^{G,S}+\mathcal{L}^{G,T}$
and $\mathcal{L}^{\mathcal{D}}$ is incorporated into the model (Method
2). The transportation loss $\mathcal{L}^{t}$ contributes to the
model's performance, resulting in improvements of 4.8\% on \emph{Office-31}
and 1.7\% on \emph{ImageCLEF-DA} (Method 3). Additionally, the effectiveness
of $\mathcal{L}^{HMM}$ leads to further improvements of around 2\%
on both datasets (the difference between Method 4 to Method 7). It
is noteworthy that our model achieves the best performance when all
component losses are activated and participate in the training process.

\begin{figure}
\centering{}\includegraphics[width=0.33\textwidth]{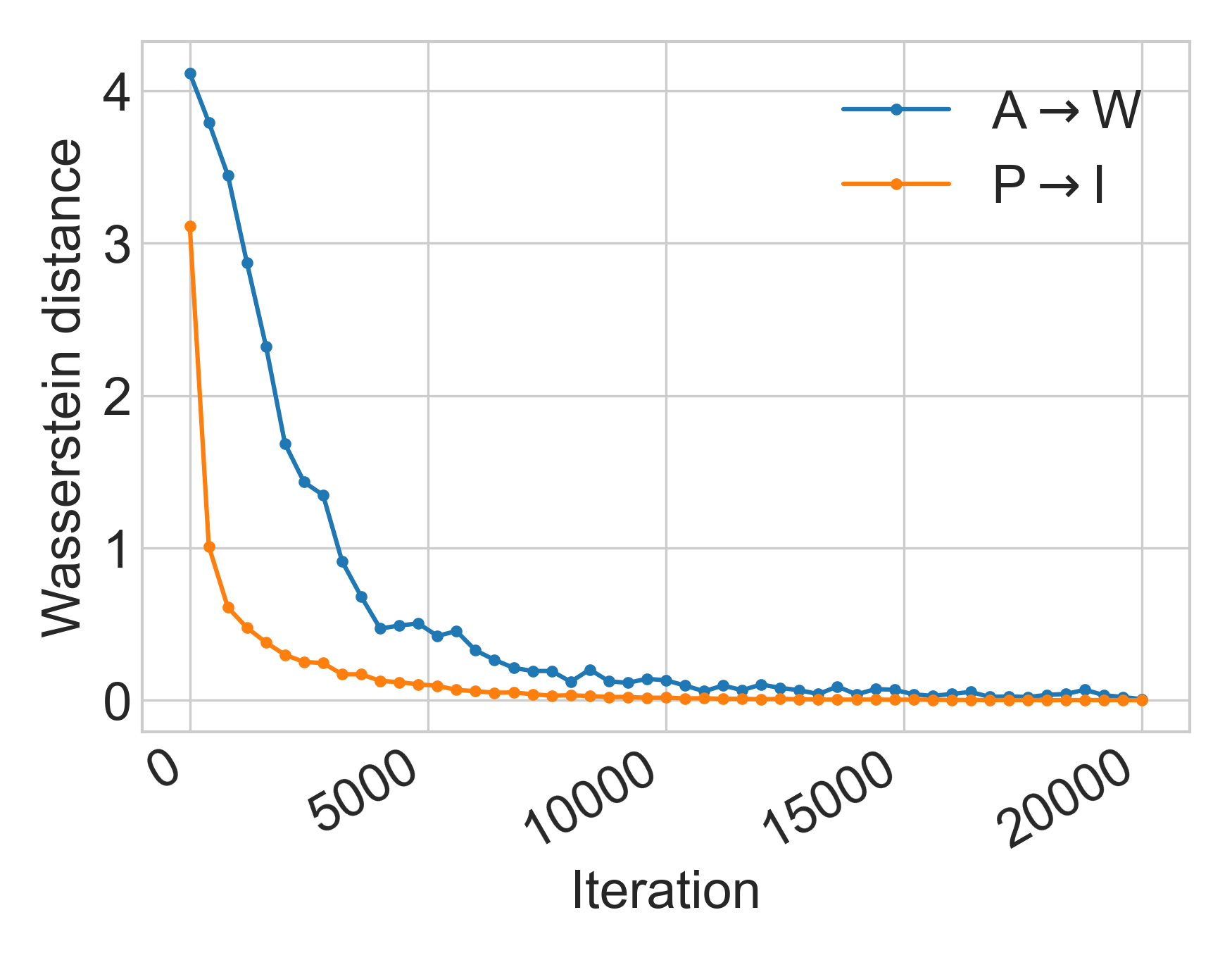}\caption{$\mathcal{W}_{c,\protect\bpi}\left(\mathbb{Q},\mathcal{Q}^{S}\right)$
during the training on transfer tasks \textbf{A}$\rightarrow$\textbf{W}
and \textbf{P}$\rightarrow$\textbf{I}.\label{fig:abla_WS}}
\end{figure}

\begin{figure}
\subfloat[Binary discriminator (Accuracy: $80.5\%$).\label{fig:abla_binD}]{\begin{centering}
\includegraphics[width=0.45\columnwidth]{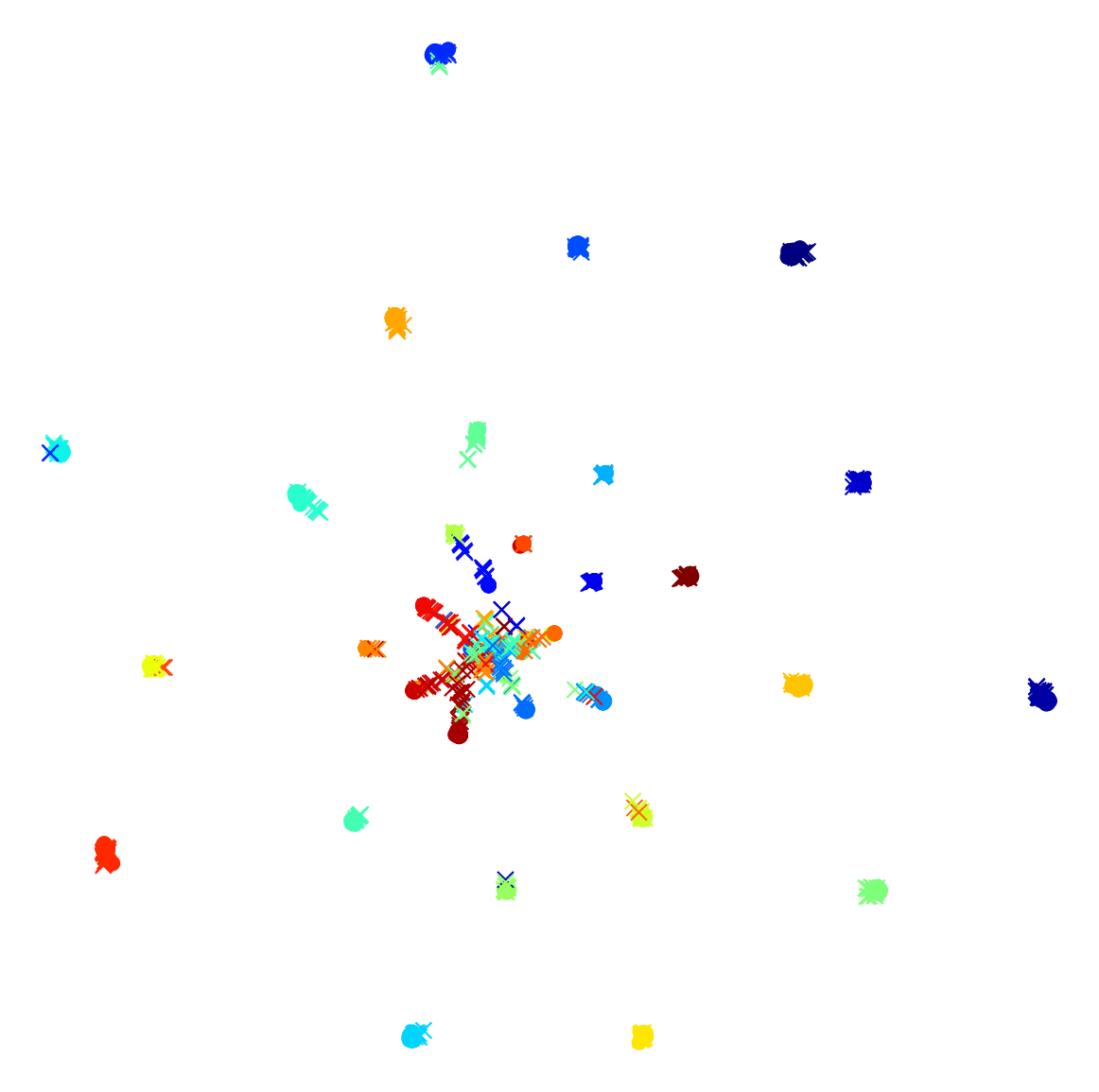}
\par\end{centering}
}\hfill{}\subfloat[Multi-class discriminator (Accuracy: $86.9\%$).\label{fig:abla_multiD}]{\centering{}\includegraphics[width=0.45\columnwidth]{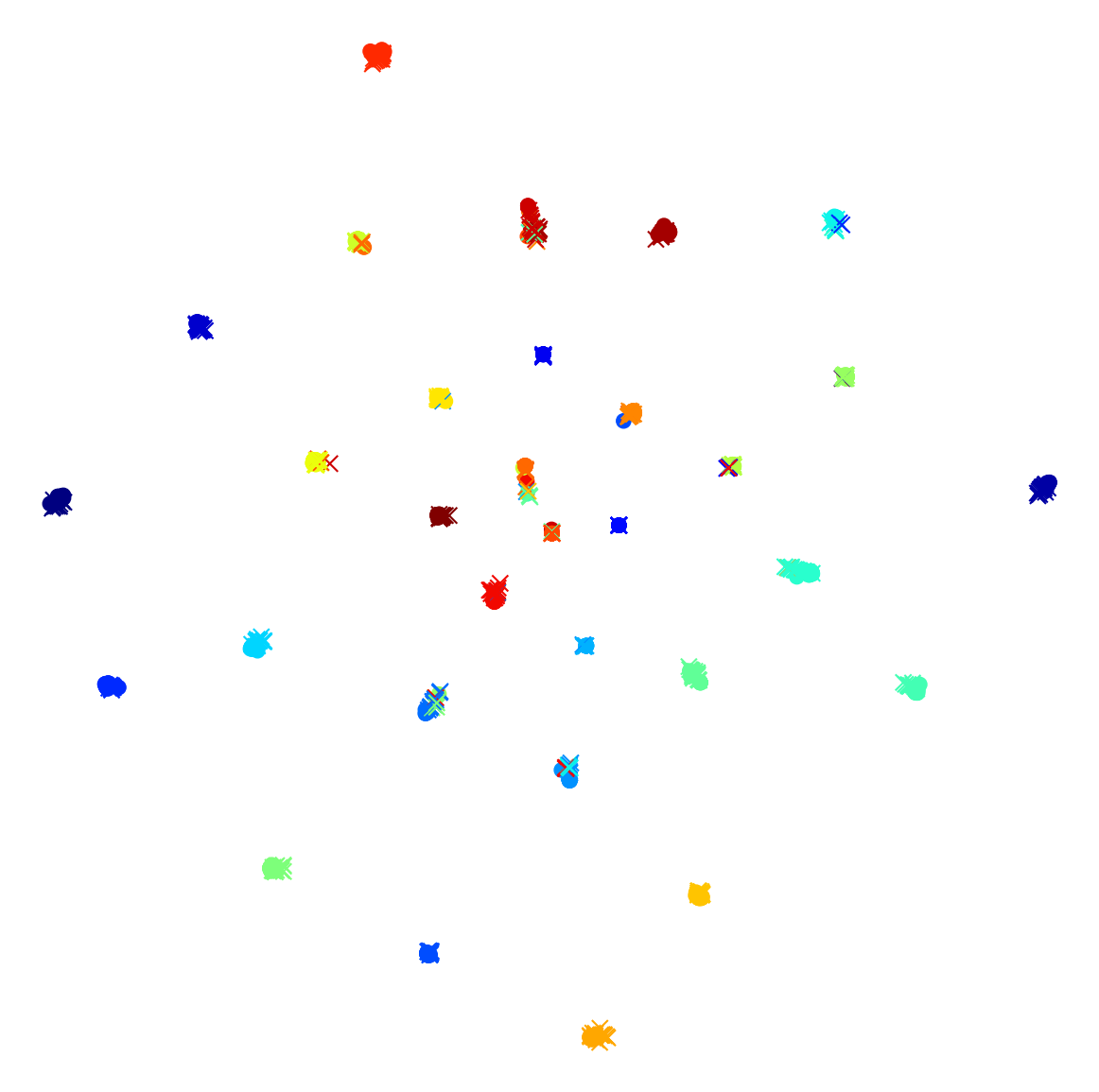}}\caption{The \emph{t}-SNE visualization with different scenarios of discriminator
$\mathcal{D}$ on the transfer task \textbf{A}\textrightarrow \textbf{D}.
Each color represents a class, while the circle and cross markers
indicate the source and target data, respectively.\label{fig:abla_D}}

\end{figure}

\begin{figure}
\centering{}\includegraphics[width=0.28\textwidth]{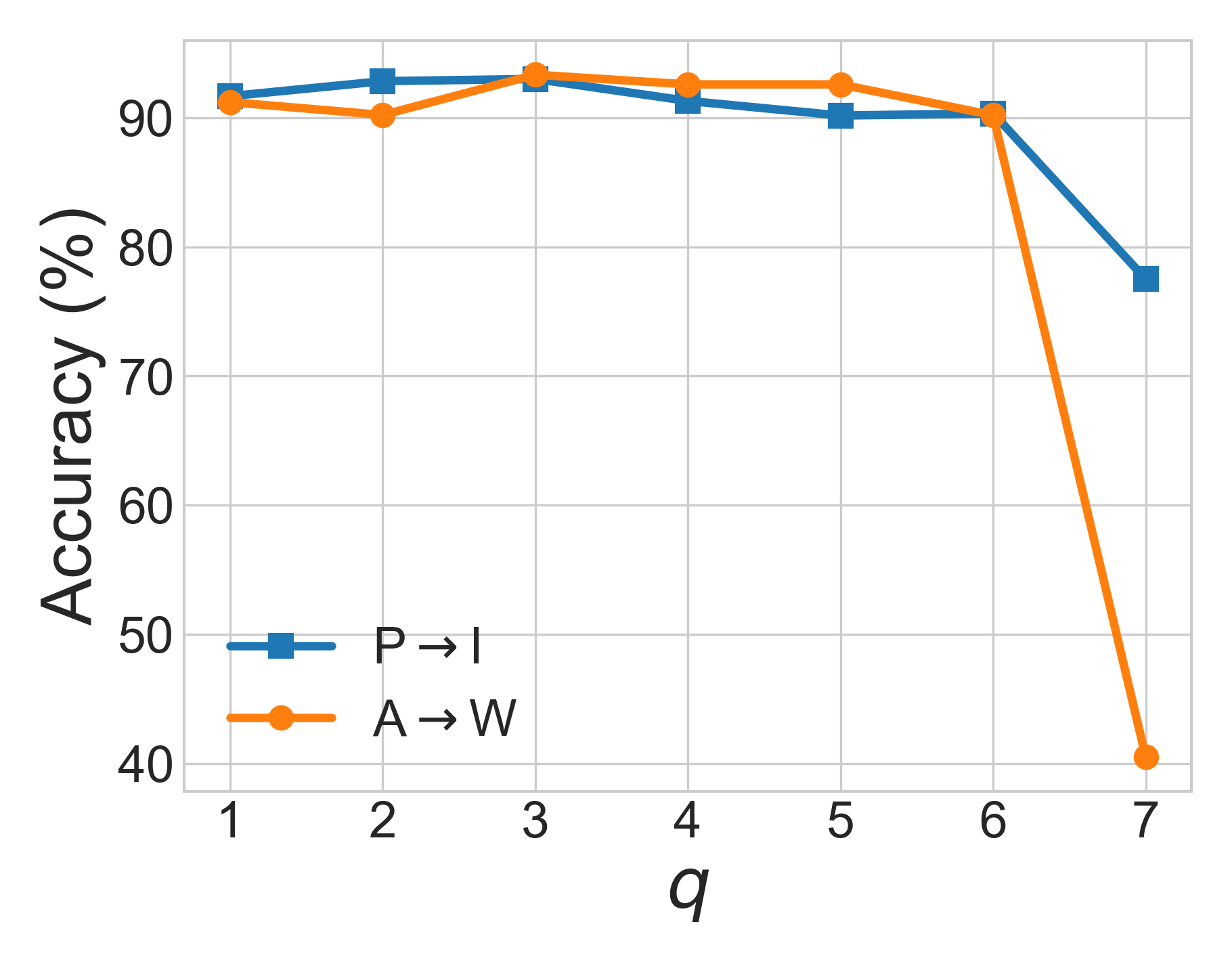}\caption{Analysis of twisting the $q$-order on transfer tasks \textbf{A}$\rightarrow$\textbf{W}
(Office-31) and \textbf{P}$\rightarrow$\textbf{I} (ImageCLEF-DA).\label{fig:abla_change_q}}
\end{figure}

\subsubsection{Wasserstein Distance}

We further investigate $\mathcal{W}_{c,\bpi}\left(\mathbb{Q},\mathcal{Q}^{S}\right)$
in Eq. (\ref{eq:OT_Q}) by observing values of $\mathcal{L}^{t}$.
Figure \ref{fig:abla_WS} shows that WS distances on both transfer
tasks \textbf{A}$\rightarrow$\textbf{W} and \textbf{P}$\rightarrow$\textbf{I}
decrease during the training process, indicating that the data and
label shifts between the source and target domains are being mitigated
and aligning with our intuition.

\subsubsection{Effect of Multi-class Discriminator}

We conducted a comparison between the multi-class discriminator $\mathcal{D}$
and a binary discriminator (similar to the GAN objective function)
to assess the effectiveness of $\mathcal{D}$. We performed \textit{t}-SNE
visualization and recorded the test accuracy after training. In one
case, the objective function included only $\mathcal{L}^{C}$ along
with a minimax loss involving the generator $G$ and either the binary
discriminator (Figure \ref{fig:abla_binD}) or the multi-class discriminator
(Figure \ref{fig:abla_multiD}). The results indicate that the binary
discriminator aims to mix the source and target samples without considering
source label information, whereas the multi-class discriminator $\mathcal{D}$
pushes target samples to the source class region and exhibits clear
boundaries. As a result, the classification accuracy significantly
increases from $80.5\%$ to $86.9\%$.

\subsubsection{Effect of Class-aware Higher-Order Moment Matching}

\begin{table}[H]
\centering{}\caption{Result (\%) of our proposed CLOTH with and without CaHoMM on ImageCLEF-DA.\label{tab:CAHoMM_effect}}
\resizebox{1.0\columnwidth}{!}{%%
\begin{tabular}{cccccccc}
\hline 
Method  & I$\rightarrow$P  & P$\rightarrow$I  & I$\rightarrow$C  & C$\rightarrow$I  & C$\rightarrow$P  & P$\rightarrow$C  & Avg\tabularnewline
\hline 
CLOT only  & 80.7  & 94.2  & 96.7  & 94.2  & 77.3  & 93.3  & 89.4\tabularnewline
CLOT$+$HoMM \cite{chen2020homm}  & 80.5  & 94.3  & 96.7  & 93.8  & 80.5  & \textbf{97.0}  & 90.5\tabularnewline
\textbf{CLOTH}  & \textbf{83.2}  & \textbf{95.0}  & \textbf{97.5}  & \textbf{95.8}  & \textbf{80.7}  & 96.7  & \textbf{91.5}\tabularnewline
\hline 
\end{tabular}} 
\end{table}

We conducted experiments to evaluate the effectiveness of our proposed
Class-aware Higher-order Moment Matching (CaHoMM) method, as described
in Section \ref{sec:CaHoMM}, in three different scenarios: (i) \emph{CLOT
only}, where we trained our CLOTH without CaHoMM by removing the loss
$\mathcal{L}^{HMM}$ from the final objective function (22); (ii)
CLOT with the HoMM method \cite{chen2020homm}; and (iii) \emph{CLOTH},
our proposed method that integrates CaHoMM. The results, presented
in Table \ref{tab:CAHoMM_effect}, demonstrate that our \emph{CLOTH}
(fourth row) achieves a notable improvement of $2.1\%$ compared to
\emph{CLOT only} (second row) and a $1\%$ improvement compared to
CLOT with the HoMM method. This improvement is attributed to the effectiveness
of CaHoMM, which focuses on matching complex distributions between
the source and target domains while considering the label information
on the source domain. By leveraging the label information, CaHoMM
enhances the alignment of distributions in a class-aware manner, leading
to improved domain adaptation performance.

\subsubsection{Analysis of Different Order Moment Matching}

In this experiment, we aimed to investigate the impact of varying
the $q$-order on the model performance. We set $q$ in the range
of $\left\{ 1,2,3,4,5,6,7\right\} $ and recorded the test accuracy
after training on transfer tasks \textbf{A}$\rightarrow$\textbf{W}
and \textbf{P}$\rightarrow$\textbf{I}, as described in Figure \ref{fig:abla_change_q}.
The results demonstrated that the performance remained stable with
$q$ ranging from $1$ to $6$, with the best performance achieved
at $q=3$. However, when using higher-order moment matching ($q\geq7$),
the performance significantly dropped. This drop in performance could
be attributed to the limitations of small batch sizes when approximating
the Higher-Order Moment Matching distance based on mini-batches \cite{raudys1991small}.
Therefore, in our proposed method, we found that using a $q$-order
of 3 yielded the most effective results.

%% file: background.tex
%% LyX 2.3.6.2 created this file.  For more info, see http://www.lyx.org/.
%% Do not edit unless you really know what you are doing.
%\documentclass[english]{article}
%\usepackage[T1]{fontenc}
%\usepackage[latin9]{inputenc}
%\usepackage{amssymb}
%\usepackage{babel}
%\begin{document}
In what follows, we present the background of OT for two discrete
distributions. Consider two discrete distributions: $\mathbb{P}^{1}=\sum_{i=1}^{M}\pi_{i}^{1}\delta_{\bx_{i}^{1}}$
and $\mathbb{P}^{2}=\sum_{j=1}^{N}\pi_{j}^{2}\delta_{\bx_{j}^{2}}$
where $\bpi^{1}=\left[\pi_{i}^{1}\right]_{i=1}^{M}$ and $\bpi^{2}=\left[\pi_{j}^{2}\right]_{j=1}^{N}$
are probability masses, $\left\{ \bx_{i}^{1}\right\} _{i=1}^{M}$
and $\left\{ \bx_{j}^{2}\right\} _{j=1}^{N}$ are the sets of atoms,
and $\delta_{\bx}$ is the Dirac delta distribution concentrated at
$\bx$. Let $c\left(\bx_{i}^{1},\bx_{j}^{2}\right)$ be a cost function.
The OT distance between $\mathbb{P}^{1}$ and $\mathbb{P}^{2}$ w.r.t.
the cost function $c$ is defined as 
\begin{equation}
\min_{A\in\mathbb{R}_{+}^{M\times N}}\sum_{i=1}^{M}\sum_{j=1}^{N}a_{ij}c\left(\bx_{i}^{1},\bx_{j}^{2}\right),\label{eq:total_cost}
\end{equation}
where $A=\left[a_{ij}\right]\in\mathbb{R}_{+}^{M\times N}$ of non-negative
elements satisfying $\sum_{j=1}^{N}a_{ij}=\pi_{i}^{1},\forall i\in\left\{ 1,...,M\right\} $
and $\sum_{i=1}^{M}a_{ij}=\pi_{j}^{2},\forall j\in\left\{ 1,...,N\right\} $.

In addition, $a_{ij}\in\left[0;1\right]$ is interpreted as the probability
to match $\bx_{i}^{1}$ and $\bx_{j}^{2}$ or to transport $\bx_{i}^{1}$
to $\bx_{j}^{2}$, which suffers the cost $c\left(\bx_{i}^{1},\bx_{j}^{2}\right)$.
Therefore, the sum $\sum_{i=1}^{M}\sum_{j=1}^{N}a_{ij}c\left(\bx_{i}^{1},\bx_{j}^{2}\right)$
can be viewed as the total cost to match $\mathbb{P}^{1}$ and $\mathbb{P}^{2}$
or to transport $\mathbb{P}^{1}$ to $\mathbb{P}^{2}$. %\end{document}